\documentclass{article} % For LaTeX2e
\usepackage{iclr2024_conference,times}

\usepackage{amsmath,amsfonts,bm}

\def\eqref#1{equation~\ref{#1}}
\def\1{\bm{1}}

\DeclareMathAlphabet{\mathsfit}{\encodingdefault}{\sfdefault}{m}{sl}
\SetMathAlphabet{\mathsfit}{bold}{\encodingdefault}{\sfdefault}{bx}{n}

\usepackage{hyperref}
\usepackage{url}

\iclrfinalcopy

\usepackage{microtype}
\usepackage{graphicx}
\usepackage{tabularx}
\usepackage{booktabs}
\usepackage{multicol}

\usepackage{amsmath}
\usepackage{amssymb}
\usepackage{mathtools}
\usepackage{amsthm}
\usepackage{algorithm}
\usepackage{algorithmic}
\usepackage{subcaption}

\usepackage{multirow}
\usepackage{bm}
\usepackage{colortbl}
\usepackage{wrapfig}
\usepackage{enumitem}
\usepackage{soul}
\usepackage{makecell}
\usepackage{thm-restate}

\usepackage{amsmath,stmaryrd,graphicx}

\usepackage[capitalize,noabbrev, nameinlink]{cleveref}

\makeatletter
\newcommand{\fixed@sra}{$\vrule height 2\fontdimen22\textfont2 width 0pt\rightarrow$}
\newcommand{\shortarrow}[1]{%
  \mathrel{\text{\rotatebox[origin=c]{\numexpr#1*45}{\fixed@sra}}}
}
\makeatother

\definecolor{grn}{rgb}{0.1, 0.6, 0.1}
\definecolor{mgt}{rgb}{0.7, 0.3, 0.7}
\definecolor{chamoisee}{rgb}{0.63, 0.47, 0.35}
\definecolor{purp}{rgb}{0.65, 0.16, 0.65}
\definecolor{alizarin}{rgb}{0.82, 0.1, 0.26}
\definecolor{azure(colorwheel)}{rgb}{0.0, 0.5, 1.0}
\definecolor{brown}{rgb}{0.65, 0.16, 0.16}
\definecolor{lblue}{rgb}{0, 0.2, 0.8}
\definecolor{orange}{rgb}{1.0, 0.5, 0.0}
\newcommand{\sungs}[1]{{\color{alizarin}{#1}}}

\definecolor{myblue}{RGB}{0, 0, 255} % main highlighting color
\definecolor{mydarkblue}{RGB}{0, 0, 139} % darker hover color
\hypersetup{
    colorlinks=true,
    linkcolor=mydarkblue,
    filecolor=mydarkblue,
    citecolor=mydarkblue,
    urlcolor=mydarkblue,
}

\theoremstyle{plain}
\newtheorem{theorem}{Theorem}

\newtheorem{toyexample}[theorem]{Example}
\theoremstyle{definition}

\theoremstyle{remark}

\title{Learning Energy Decompositions for \\ Partial Inference of GFlowNets}
\author{Hyosoon Jang$^{1}$, Minsu Kim$^{2}$, Sungsoo Ahn$^{1}$ \\
$^{1}$POSTECH \qquad $^{2}$KAIST \\
\texttt{\{hsjang1205,sungsoo.ahn\}@postech.ac.kr, min-su@kaist.ac.kr}
}

\begin{document}

\maketitle

\begin{abstract}
This paper studies generative flow networks (GFlowNets) to sample objects from the Boltzmann energy distribution via a sequence of actions. In particular, we focus on improving GFlowNet with \textit{partial inference}: training flow functions with the evaluation of the intermediate states or transitions. To this end, the recently developed forward-looking GFlowNet reparameterizes the flow functions based on evaluating the energy of intermediate states. However, such an evaluation of intermediate energies may (i) be too expensive or impossible to evaluate and (ii)~even provide misleading training signals under large energy fluctuations along the sequence of actions. To resolve this issue, we propose learning energy decompositions for GFlowNets (LED-GFN). Our main idea is to (i) decompose the energy of an object into learnable potential functions defined on state transitions and (ii) reparameterize the flow functions using the potential functions. In particular, to produce informative local credits, we propose to regularize the potential to change smoothly over the sequence of actions. It is also noteworthy that training GFlowNet with our learned potential can preserve the optimal policy. We empirically verify the superiority of LED-GFN in five problems including the generation of unstructured and maximum independent sets, molecular graphs, and RNA sequences.
\end{abstract}

\section{Introduction}
\label{sec:intro}

Generative Flow Networks \citep[GFlowNets or GFNs]{bengio2021flow} are frameworks to sample objects through a sequence of actions, e.g., iteratively adding nodes to a graph. Their key concept is training a policy that sequentially selects the actions to sample the object from the Boltzmann distribution \citep{boltzmann1868studien}. Such concepts enable discovering diverse samples with low energies, i.e., high scores, as an alternative to reinforcement learning (RL)-based methods which tend to maximize the return of the sampled object \citep{silver2016mastering,sutton2018reinforcement}.

To sample from the Boltzmann distribution, GFlowNet trains the policy to assign action selection probability based on energy of terminal state \citep{bengio2021flow,bengio2021gflownet,malkin2022trajectory}, e.g., a high probability to the action responsible for the low terminal energy. However, such training has fundamental limitations in credit assignment, as it is hard to identify the action responsible for terminal energy \citep{pan2023better}. This limitation stems from solely relying on the terminal energy associated with multiple actions, lacking the information to identify the contribution of individual actions, akin to challenges in RL with sparse reward \citep{arjona2019rudder,ren2022learning}.

An attractive paradigm to tackle this issue is \textit{partial inference} \citep{pan2023better} that trains flow functions with local credits, e.g., evaluation of the intermediate states or transitions. Such local credit identifies individual action contributions to the terminal energy before reaching the terminal state. To this end, \citet{pan2023better} proposed a forward-looking GFlowNet (FL-GFN), which assigns the local credit based on the energy of incomplete objects associated with intermediate states.

However, FL-GFN crucially relies on two assumptions that may not hold in practice. First, FL-GFN requires evaluating the energy of the intermediate state in the trajectories. However, the energy function can be expensive or even impossible to evaluate. Next, FL-GFN assumes the energy of intermediate states to provide useful hints for the terminal energy. However, this may not be true when the intermediate energy largely fluctuates along the sequence of states, e.g., low intermediate energies may lead to a terminal state with high energy. We illustrate such a pitfall in \Cref{fig:first_fig}.

%\sungs{However, this may not be true when the terminal energy differs significantly from the intermediate energies.}

%However, associating an object with an incomplete sequence of actions can be non-trivial and the

\textbf{Contribution.} We propose learning energy decomposition for GFlowNet, coined LED-GFN. Our key idea is to perform partial inference by decomposing terminal state energy into a sum of learnable potentials associated with state transitions and use them as local credits. In particular, we show how to regularize the potential function to preserve the ground-truth terminal energy and minimize variance over the trajectory to yield informative potentials. \Cref{fig:first_fig} highlights how LED-GFN provides informative local credits compared to the existing approach.

%In particular, we show how to constrain the potential functions to preserve the ground-truth object energy and minimize the variance to yield informative potentials. \Cref{fig:first_fig} highlights how LED-GFN provides informative local credit compared to the existing approach.

% Our key idea is to train potential functions that decompose terminal energies by summation over state transitions. In particular, we constraint on the potential functions to preserve the optimal policy and regularize the potential function to yield better local credit signal

%\textbf{Contribution.} To evaluate intermediate state energy, we propose learning energy decomposition for FL-GFlowNet called LED-GFN. Our key idea is to train models that decompose terminal energy into multiple gains within sequential actions, and inject constraints to prevent producing sub-optimal proxy energy gains, e.g., sparse proxy energy gains. Then, we incorporate decomposed energy gain to train FL-GFlowNet. \Cref{fig:motiv_illust} highlights how LED-GFN provides more informative local credit signal compared to existing approach.
\newcommand{\swmo}[1]{{\color{Tdgreen}{#1}}}
\newcommand{\hsja}[1]{\textcolor{lblue}{#1}}

\begin{figure}[t]
\centering
  \vspace{-.1in}
\begin{subfigure}[t]{0.85\linewidth}
\centering
  \includegraphics[width=\linewidth]{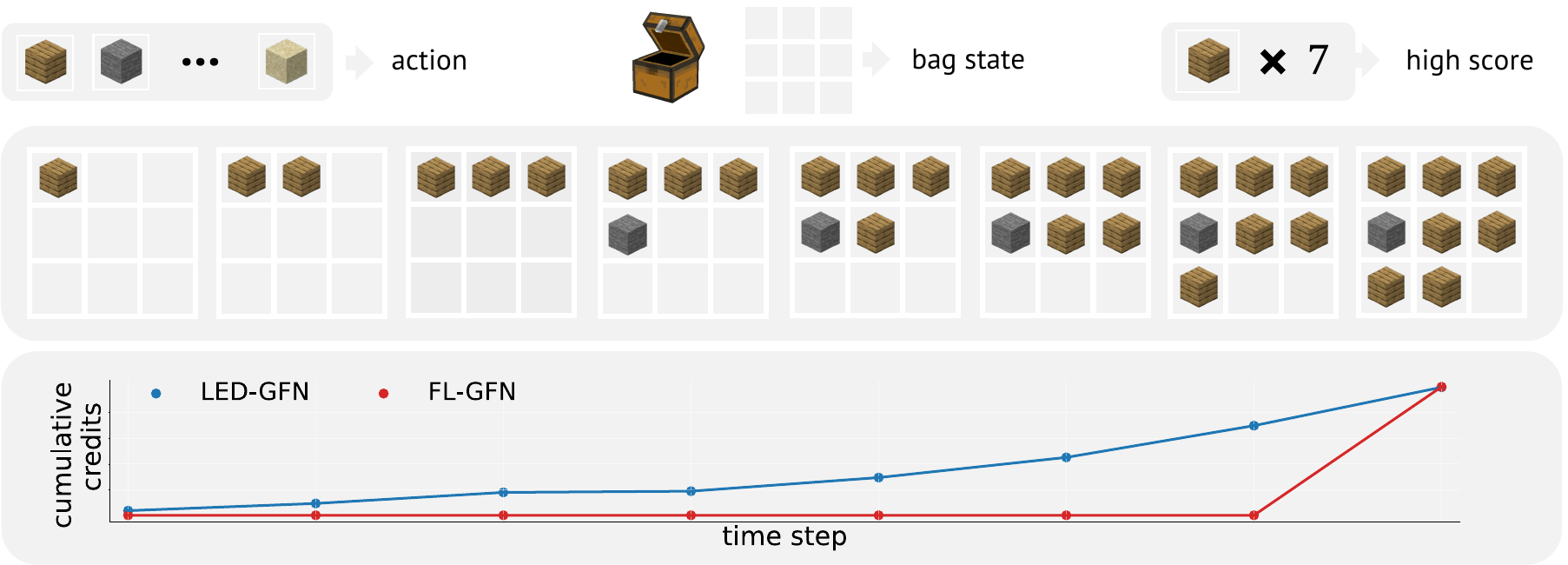}
  \vspace{-.14in}
\end{subfigure} \caption{ \textbf{The local credit evaluation in bag generation (\cref{toy:toytoy})}. \textbf{(first row)} The task is to generate a bag of entities, where the seven same entities yields a high score. (\textbf{second row}) Left-to-right indicates the state transitions over a given trajectory. \textbf{(third row)} The \sungs{energy-based evaluation} fails to produce informative local credits since every intermediate state has zero energy, whereas our \hsja{potential function} produces informative credits by enforcing the potentials to be uniformly distributed.}\label{fig:first_fig}
  \vspace{-.13in}
\end{figure}

To be specific, our energy decomposition framework resembles the least square-based return decomposition for episodic reinforcement learning \citep{efroni2021reinforcement,ren2022learning}. We parameterize potential function with a regression model that is constrained to be equal to the terminal energy when aggregated over the entire action sequence. We also regularize the potential function to minimize the variance along the trajectory, so that the potential function provides dense local credits in training GFlowNet. Such potentials associated with intermediate transitions provide informative signals, as each of them is enforced to be correlated with the terminal energies. The training of potential function is online, which uses samples collected during GFlowNet training. 

% Note that 

% Note that such potentials provide informative hints for the terminal energy

We extensively validate LED-GFN on various tasks: set generation \citep{pan2023better}, bag generation \citep{towardsunderstandinggflownets}, molecular discovery \citep{bengio2021gflownet}, RNA sequence generation \citep{jain2022biological}, and the maximum independent set problem \citep{zhang2023let}. We observe that LED-GFN (1) outperforms FL-GFN when the assumption of intermediate energy does not hold, (2) excels in practical domains compared to GFlowNets and RL-based baselines, and (3) achieves similar performance to FL-GFN even when intermediate energy provides the ``ideal'' local credit.

\section{Preliminaries}
\label{sec:preliminaries}

In this section, we describe generative flow networks \citep[GFlowNets or GFNs]{bengio2021flow} and their partial inference algorithm. We describe additional related works in \Cref{appx:related_work}.

\subsection{GFlowNets}
%\subsection{Generative Flow Networks}

GFlowNets sample from discrete space $\mathcal{X}$ through a sequence of actions from the action space $\mathcal{A}$ that make transitions in the state space $\mathcal{S}$. For each complete trajectory $\tau=(s_0, s_1, \ldots , s_T)$, the terminal state is the object $x = s_{T} \in \mathcal{X}$ to be generated. The state transitions are determined by the action sequence $(a_1, \ldots a_{T-1})$, e.g., $a_{t}$ determines $s_{t} \rightarrow s_{t+1}$. The policy $P_{F}(s'|s)$ selects the action $a$ to transition from the current state $s$ to the next state $s'$ and induces a distribution over the object $x$.

The main objective of GFlowNet is to train the policy $P_{F}(\cdot|\cdot)$ that samples objects from the Boltzmann distribution with respect to a given energy function $\mathcal{E}:\mathcal{X}\rightarrow \mathbb{R}$ as follows:
\begin{equation} \label{equation:sampling_condition}
    P^{\top}_{F}(x)\propto \exp(-\mathcal{E}(x)),
\end{equation}
where $P^{\top}_F(x)$ is the distribution of sampling an object $x$ induced from marginalizing over the trajectories conditioned on $x=s_{T}$. We omit the temperature for simplicity. To this end, GFlowNet trains with auxiliary objectives based on state transition, trajectory, or sub-trajectory information.  

\textbf{Detailed balance \citep[DB]{bengio2021gflownet}.} The DB utilizes the experience of state transitions to train GFlowNet. It trains the GFlowNet with a forward policy model $P_{F}(s' | s)$, a backward policy $P_B(s | s')$, and a state flow estimator $F(\cdot) : \mathcal{S} \rightarrow \mathbb{R}^+$ by minimizing the following loss function:
\begin{equation*}
\mathcal{L}_{\text{DB}}(s, s') = \left(\log F(s)+ \log P_{F}(s' | s) - \log F(s')-P_B(s | s')\right)^2,
\end{equation*}
where the flow $F(s)$ of the terminal state $s_{T}=x$ is defined to be identical to the energy $\exp(-\mathcal{E}(x))$. 

%However, it can suffer from inefficient credit assignment in flow estimation. The true energy signal is provided at only terminal states, which may be delayed in training policy on intermediate states due to the intermediate flow estimation errors.

\textbf{Trajectory balance \citep[TB]{malkin2022trajectory}.} The TB aims to learn the policy faster by training on full trajectories. To this end, TB requires a forward policy model $P_{F}(s' | s)$, a backward policy $P_B(s | s')$, and a learnable scalar $Z$ to minimize the following loss function:
\begin{equation*}
\mathcal{L}_{\text{TB}}=\left(\log Z +\sum^{T-1}_{t=0}\log P_{F}(s_{t+1}|s_{t})-\mathcal{E}(x)-\sum^{T-1}_{t=0}\log P_B(s_{t}|s_{t+1})\right)^2.
\end{equation*}
This objective is resilient to the bias from inaccurate flow estimator $F(\cdot)$ used in DB, since it directly propagates the terminal energy to train on intermediate states. However, the TB suffers from the high variance of the objective over the collected trajectories \citep{malkin2022trajectory}. %by directly propagating energy of object to train the policy on intermediate states within trajectory

\textbf{Sub-trajectory balance \citep[subTB]{madan2023learning}.} The subTB trains forward and backward policies $P_{F}(s' | s), P_{B}(s'|s)$ and a learnable scalar $Z$ similar to the TB. However, it trains on flexible length of sub-trajectory $s_{U}\rightarrow s_{U+1} \cdots\rightarrow s_{U+L}$ to minimize the following loss function:
\begin{equation*}
\mathcal{L}_{\text{subTB}}=\left(\log F(s_U) +\sum^{U+L-1}_{t=U}\log P_{F}(s_{t+1}|s_{t})-\log F(s_{U+L})-\sum^{U+L-1}_{t=U}\log P_B(s_{t}|s_{t+1})\right)^2.
\end{equation*}
This objective enables controlling the bias-variance trade-off by interpolating between the DB that trains on a single state transition and the TB that trains on a complete trajectory.

%Here, the bias-variance trade-off can be controlled by adjusting the length of the sub-trajectory, by interpolating DB (when sub-trajectory length is one) and TB (when sub-trajectory is complete trajectory) \citep{madan2023learning}. 

\subsection{Partial inference for GFlowNets}

The GFlowNet training is often challenged by limitations in credit assignment, i.e., identification and promotion of the action responsible for the observed low energy. This limitation stems from relying solely on the terminal state energy as the training signal. The terminal energy lacks information to identify the contribution of individual action, akin to how reinforcement learning with sparse reward suffers from credit assignment \citep{arjona2019rudder,ren2022learning}.

Partial inference is a promising paradigm to resolve this issue by incorporating local credits \citep{pan2023better}. Specifically, the partial inference aims to evaluate individual transitions or sub-trajectories, i.e., local credits, and provide informative training signals for identifying the specific contributions of actions. To this end, \citet{pan2023better} proposed Forward-Looking GFlowNet (FL-GFN), which evaluates intermediate state energy as a local credit signal for partial inference.

\textbf{Forward-Looking GFlowNet \citep[FL-GFN]{pan2023better}.} To enable partial inference, the FL-GFN defines a new training objective that incorporates an energy function $\mathcal{E}:\mathcal{X}\rightarrow \mathbb{R}$ for intermediate states. To this specific, FL-GFN modifies the DB as follows:
%To enable partial inference, the FL-DB objective incorporates an energy function $\mathcal{E}:\mathcal{X}\rightarrow \mathbb{R}$ for intermediate states to better assign credits to promising actions. To this specific, the FL-DB objective is defined as follows:
\begin{align}\label{eq:fldb}
\mathcal{L}_{\text{FL}}(s,s')=(\log \tilde{F}(s) + \log P_{F}(s' | s) - \mathcal{E}(s) + \mathcal{E}(s') - \log \tilde{F}(s') - \log P_B(s | s'))^2,
\end{align}
where $\tilde{F}(s)=F(s)\exp{(\mathcal{E}(s))}$ is the re-parameterized flow function and $\mathcal{E}(s')-\mathcal{E}(s)$ is the energy gain associated with the transition from $s$ to $s'$. Note that the energy function is defined only on the discrete space $\mathcal{X}$, hence FL-GFN evaluates an intermediate energy $\mathcal{E}(s)$ as the energy of an incomplete object associated with the intermediate state $s$. \citet{pan2023better} shows that optimum of \Cref{eq:fldb} induces a policy $P_{F}(\cdot|\cdot)$ that samples from Boltzmann distribution. While \cref{eq:fldb} is associated with DB, FL-GFN is also applicable to subTB with energy gains in the level of sub-trajectory.

% The FL-DB facilitates faster credit assignment by incorporating energy gain to identify contribution of action. 

%To evaluate intermediate state energy, FL-GFN extends a energy function designed for terminal state $\mathcal{E}:\mathcal{X}\rightarrow \mathbb{R}$ to evaluate energy for intermediate states.

%\textbf{Energy evaluation of intermediate states.}  %In domain-specific tasks, \citet{zhang2023let} evaluate energy for intermediate states using a hand-crafted function, which is designed with the domain knowledge.

%\input{2_2_related}

\section{Learning energy decomposition for GFlowNet}
\label{sec:method}
While the FL-GFN is equipped with partial inference capabilities, it relies on the energy function to assign local credits, which can be expensive to evaluate, or lead to sub-optimal training signals (details are described in \Cref{subsec:limitandlearn}). In this paper, we propose learning energy decomposition for GFlowNets (LED-GFN) to achieve better partial inference. In what follows, we describe our motivation for better partial inference (\Cref{subsec:limitandlearn}) and the newly proposed LED-GFN (\Cref{subsec:learning}).

\newpage

\begin{figure}[t]
\centering
\begin{subfigure}[t]{0.92\linewidth}
\centering
  \includegraphics[width=\linewidth]{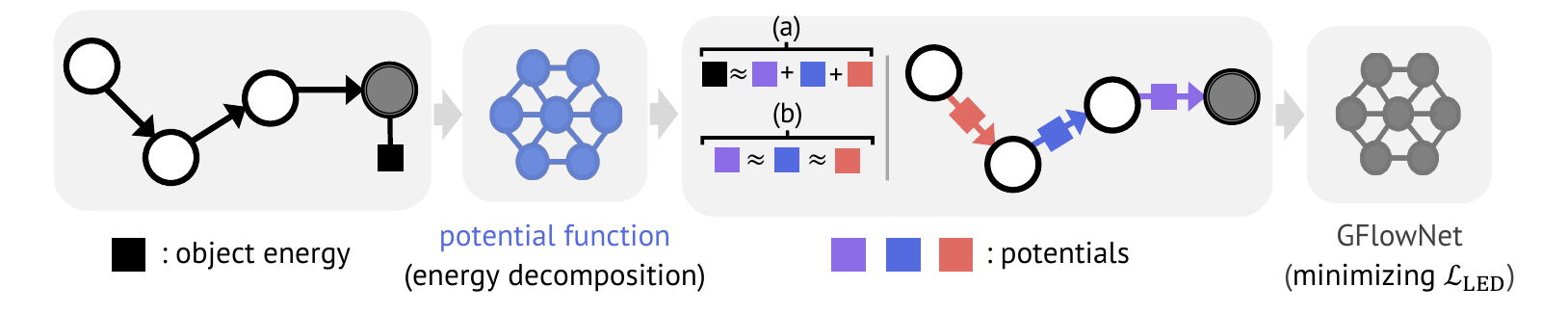}
\end{subfigure}  \vspace{-.07in}\caption{\textbf{Illustration of energy decomposition for partial inference of GFlowNet.} LED-GFN enables partial inference with potentials which (a) approximate the object energy via summation, and (b) minimize variance along the action sequence.} \label{fig:method}
\vspace{-.15in}
\end{figure}

\subsection{Motivation for better partial inference} \label{subsec:limitandlearn}

\begin{wrapfigure}{r}{0.33\textwidth}
    \vspace{-.2in}
\centering
    \includegraphics[width=0.9\linewidth]{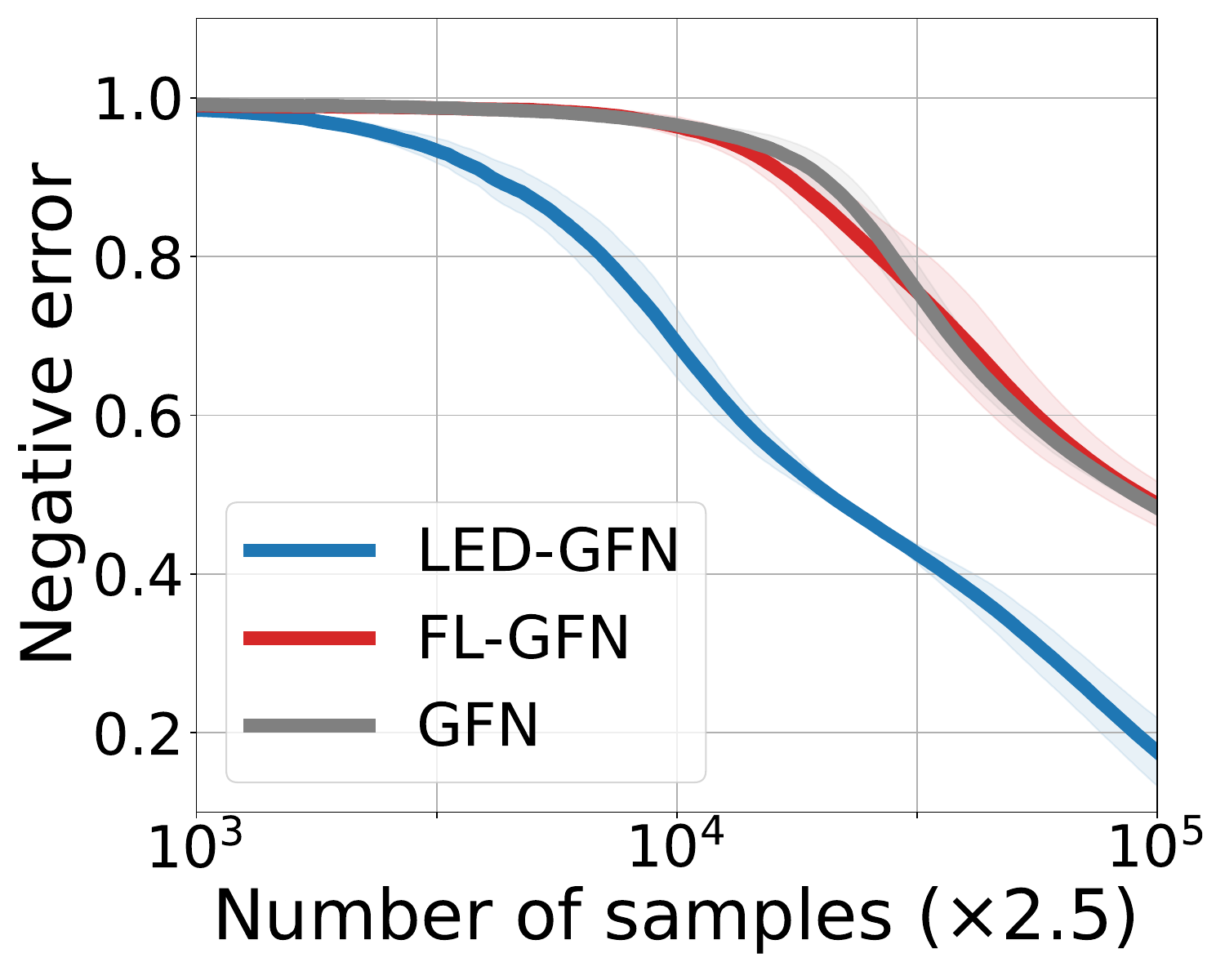}\vspace{-.05in}
    \caption{{Negative relative mean error ($\downarrow$) for estimating the true Boltzmann distribution on \Cref{toy:toytoy}-type task (Bag).} 
    %The FL-GFN fails to make improvement.
    }\label{fig:motivation}
    \vspace{-.2in}
\end{wrapfigure}

Our motivation stems from the limitations of FL-GFN, which performs partial inference based on evaluating the energies of intermediate states with respect to a single transition (for DB) or sub-trajectory (for subTB). In particular, we are inspired by how the energy gain may yield a sub-optimal local credit signal due to the following pitfalls (see \Cref{fig:first_fig,fig:oracle_calls}). 

\begin{enumerate}[topsep=-1.0pt,itemsep=1.0pt,leftmargin=5.5mm]
\item[\textbf{A.}] The energy evaluation for an incomplete object, i.e., intermediate state, can be non-trivial. In addition, the cost of energy evaluation can be expensive, which can bottleneck efficient training when called for all visited states. % The cost of energy evaluation can be expensive, which can bottleneck efficient training when called for all visited states. 
\item[\textbf{B.}] The energy can exhibit sparse or high variance on intermediate states within a trajectory, even returning zero for most states, which is non-informative for partial inference.
\end{enumerate}

We further provide a concrete example for the pitfall \textbf{B.} of FL-GFN, which is illustrated in \Cref{fig:first_fig} and \Cref{fig:motivation}\footnote{The detailed experiment settings are described in \Cref{subsec:experiment_bag}.} for both conceptual and empirical purposes, respectively:
\begin{toyexample} \label{toy:toytoy}
Consider adding objects from $\{A,B,C,D,E\}$ to a bag with a maximum capacity of nine. Define the energy as $-1$ when the bag contains seven identical objects and $0$ otherwise. 
\end{toyexample}
For \cref{toy:toytoy}, the intermediate energy (which is always $0$) does not provide information for the terminal energy. However, the number of the most frequent elements in the bag is informative even at intermediate states since greedily increasing the number leads to the best terminal state.

%\begin{toyexample} \label{toy:toytoy}
%Consider adding objects to a bag which has a limited capacity. Define the energy as $-1$ when the bag contains seven identical objects and $0$ otherwise. 
%\end{toyexample}

Our observation in \cref{toy:toytoy} hints at the existence of a partial inference algorithm to provide better local credit signals. We aim to pursue this direction with a learning-based approach. That is, we parameterize the class of potential functions that decompose the terminal energy to provide local credit signals for GFlowNet training. Our key research direction is to understand about what conditions of the potential functions are informative for partial inference.

%To resolve it, one may consider a hand-crafted function to evaluate intermediate state energy \citep{zhang2023let}, but it requires in-depth domain knowledge.

%To provide a more motivating example of how terminal energy function-based approach fails, let's consider the following scenarios, which is illustrated in \Cref{fig:motiv_{t}llust}:
%\begin{toyexample}
%In a task where terminal energy is determined by inclusion of sub-structure, the terminal energy function only yields a non-zero value upon sub-structure completion, lacking informative guidance for training individual steps in sub-structure generation.
%\end{toyexample}
%In \Cref{fig:motivation}, we demonstrate the failure of FL-GFN in improving credit assignment. As an alternative approach, one may consider designing hand-crafted function to evaluate intermediate energy \citep{zhang2023let}, but it requires in-depth domain knowledge.

%which is equipped to yield informative signal by incorporating regularization

\subsection{Algorithm description}\label{subsec:learning}
In this section, we describe our framework, coined learning energy decomposition for GFlowNet (LED-GFN), which facilitates partial inference using learned local credit. To this end, we propose to decompose the terminal energy into learnable potentials defined on state transitions. Similar to FL-GFN, we reparameterize the flow model with local credits, i.e., potentials. In contrast to FL-GFN, we optimize the local credits to enhance partial inference by minimizing variance of potentials along the action sequence. See \Cref{fig:method} for an illustration of LED-GFN.

%Similar to FL-GFN, we assign local credits via re-parameterizing the flow model with potential functions. In contrast to FL-GFN, we optimize the reparameterization to improve partial inference by minimizing the fluctuations in potentials along the sequence of actions. 

% See Figure 2 for an illustration of DPM for structured node classification with known labels

\textbf{Training with potential functions.} To be specific, we decompose the energy function $\mathcal{E}$ associated with the terminal state into learnable potential functions $\phi_{\theta}$ as follows:
\begin{equation} \label{eq:sum_decomp}
\mathcal{E}(x)\approx \Phi_{\theta}(\tau) = \sum^{T-1}_{t=0}\phi_{\theta}(s_{t}\rightarrow s_{t+1}),
\end{equation}
where $\tau = (s_{0}, s_{1},\ldots, s_{T})$, $x=s_{T}$, and the potential functions are defined on state transition $s_{t}\rightarrow s_{t+1}$. Similar to FL-GFN, we use the potential function to train the forward and backward GFlowNet policies $P_{F}, P_{B}$ and flow model $F$ to minimize the following loss:
\begin{align}\label{eq:ldb}
\mathcal{L}_{\text{LED}}(s,s')=(\log \tilde{F}(s) + \log P_{F}(s' | s) + \phi_{\theta}(s\rightarrow s') - \log \tilde{F}(s') - \log P_B(s | s'))^2.
\end{align}
Given a sub-trajectory $(s_{0},\ldots, s_{u}=s, s_{u+1}=s')$, one can derive this objective from \Cref{eq:fldb} by replacing $\mathcal{E}(s), \mathcal{E}(s')$ with $\sum_{t=0}^{u}\phi(s_{t})$ and $\sum_{t=0}^{u+1}\phi(s_{t})$, respectively. That is, it is evident that \Cref{eq:ldb} preserves the optimal policy of GFlowNet when $\mathcal{E}(x)=\Phi_{\theta}(\tau)$ is satisfied for all trajectories $\tau$ terminating with $x$.

Our objective becomes equivalent to that of FL-GFN when $\phi_{\theta}(s\rightarrow s') = \mathcal{E}(s') - \mathcal{E}(s)$, but our key idea is to \textit{learn} the potential function $\phi_{\theta}$ instead of the energy gain which can be expensive and may exhibit sparsity or high variance, as pointed out in \cref{subsec:limitandlearn}. Note that one can also introduce an approximation error $\mathcal{E}(x) - \Phi_{\theta}(\tau)$ as an additional correction term to preserve the optimal policy of GFlowNet even when the potential function $\phi_{\theta}$ is inaccurate. In \Cref{subsec:preserving_optimal}, we describe how LED-GFN consistently induces the optimal policy that samples from the Boltzmann distribution.

\textbf{Training potentials with squared loss.} In training the potential function, our key motivation is twofold: (a) accurately estimating the true energy through summation and (b) providing dense and informative training signals by minimizing variance along the action sequence. 

%In particular, we can enforces a low variance of potentials within the trajectory by applying dropout to potentials in \Cref{eq:least_square} \citep{ren2022learning}. It enables to produce dense potentials, where each of them is constrained to be correlated with future object energy. Such benefit is also akin to how step-wise informative rewards improve reinforcement learning \citep{gangwani2020learning,arumugam2021information}.\footnote{In \Cref{fig:reducing_variance, fig:analysis}, we also analyze how reducing variance make informative local credits.}

To this end, given a trajectory $\tau=(s_{0},\ldots, s_{T}=x)$, we train the potential functions $\phi_{\theta}$ to minimize the loss function for (a) achieving $\mathcal{E}(x)\approx \Phi_{\theta}(\tau)$ with (b) dropout-based regularization:
\begin{equation} \label{eq:least_square}
\ell_{\text{LS}}(\tau)=\mathbb{E}_{\bm{z} \sim \operatorname{Bern}(\lambda)}\left[\left(\frac{1}{T}\mathcal{E}(s_{T})- \frac{1}{C}\sum^{T-1}_{t=0}z_t\phi_{\theta}(s_{t}\rightarrow s_{t+1})\right)^2\right].
\end{equation}
The $T$-length random vectors $\bm{z}$ promotes the dropout, where $z_t=0$ sampled from the Bernoulli distribution with probability $1-\lambda$. Dividing by $T$ and $C=\sum^{T-1}_{t=0} z_t$ aligns the scales to compensate for the scale reduction induced by dropout. When $\lambda = 1$ and the loss function is minimized, i.e., $\ell_{\text{LS}}(\tau)=0$ for all $\tau$, the potential function decomposes the energy function without error. When $\lambda < 1$, dropout prevents heavy reliance on specific potentials to satisfy \Cref{eq:sum_decomp}, thereby reducing the variance and sparsity of the potentials along the action sequence.\footnote{\citet{ren2022learning} provide a formal proof that \Cref{eq:least_square} serves as a surrogate objective to satisfy \Cref{eq:sum_decomp} while reducing the variance and sparsity of the potentials along the action sequence.} Note that our intuition is similar to recent works on learning return decomposition to alleviate sparse reward problems in reinforcement learning \citep{arjona2019rudder,gangwani2020learning,ren2022learning}. 

%When $\lambda < 1$, \citet{ren2022learning} show that \Cref{eq:least_square} provides surrogate objective to satisfy \Cref{eq:sum_decomp} while reducing the variance and sparsity of the potentials along the trajectory $\tau$, akin to how dropout benefits to train dense networks \citep{srivastava2014dropout,labach2019survey}. %,  

%As a motivating benefit, we can easily inject additional constraints in optimizing \Cref{eq:least_square} to produce informative potentials. In particular, we can enforces a low variance of potentials within the trajectory by applying dropout to potentials in \Cref{eq:least_square} \citep{ren2022learning}. It enables to produce dense potentials, where each of them is constrained to be correlated with future object energy. Such benefit is also akin to how step-wise informative rewards improve reinforcement learning \citep{gangwani2020learning,arumugam2021information}.\footnote{In \Cref{fig:reducing_variance,fig:analysis}, we also analyze how reducing variance make informative local credits.}

To train the potential function, we define its training as online learning within GFlowNet training, i.e., learning from trajectories obtained during GFlowNet training. We describe the overall algorithm in \Cref{alg:training}. Such an alternative training of the potential function and the policy is similar to model-based reinforcement learning algorithms \citep{luo2018algorithmic, sun2018dual, janner2019trust} for monotonic improvement of policies. Note that LED-GFN can also be implemented with subTB, where the details are described in \Cref{subsec:appdix_subtb}.

\begin{algorithm}[tb]
%\small
\caption{Learning energy decomposition for GFlowNet}
\label{alg:training}
\begin{algorithmic}[1]
   \STATE Initialize the buffer $\mathcal{B}$, forward and backward policy $P_F,P_B$, state flow $\tilde{F}$, and model $\phi_{\theta}$.
   \STATE Update the model $\phi_{\theta}$ to minimize $\ell_{\text{ED}}$ if the generation trajectories are given in advance.
   \REPEAT
   \STATE Sample a batch of trajectories $\{\tau_{b}\}_{b=1}^{B_{1}}$ from forward policy $P_F$.
   \STATE Update buffer $\mathcal{B}\leftarrow \mathcal{B} \cup \{\tau_{b}\}_{b=1}^{B_{1}}$.
   \FOR[Energy decomposition learning]{$n=1,\ldots,N$}
   \STATE Sample a batch of trajectories $\{\tau_{b}\}_{b=1}^{B_{2}}$ from the buffer $\mathcal{B}$.
   \STATE Update the model $\phi_{\theta}$ to minimize $\ell_{\text{LS}}$ with $\{\tau_{b}\}_{b=1}^{B_{2}}$.
   \ENDFOR 
   \STATE Compute intermediate energy gains $\phi_{\theta}(s_i,a_i)$ for all $(s_i,a_i) \in \tau$.
   \STATE Update the GFlowNet $P_F,P_B,\tilde{F}$ to minimize $\mathcal{L}_{\text{LED}}$ with $\tau$ and $\phi_{\theta}(s_i,a_i)$ for all $(s_i,a_i) \in \tau$.
   \UNTIL{converged}
\end{algorithmic}
\end{algorithm}

\section{Experiment}
\label{sec:experiment}

We extensively evaluate LED-GFN on various domains, including bag generation \citep{towardsunderstandinggflownets}, molecule generation \citep{bengio2021flow}, RNA sequence generation \citep{jain2022biological}, set generation \citep{pan2023better}, and the maximum independent set problem \citep{zhang2023let}. Following prior studies, we consider the number of modes, i.e., samples with energy lower than a specific threshold, and the average top-$100$ score as the base metrics, which are measured via samples collected during training. We report all the performance using three different random seeds.

%\input{resources/fig_set_only}

%\begin{figure}[t]
%\centering
%\includegraphics[width=0.85\linewidth]{resources/figures/sub_bag_re_inter.pdf}
%\caption{The bag generation task. The LED-GFN shows similar performance to the FL-GFlowNet, which incorporates sparse intermediate energy %based on the underlying energy gain mechanism.
%}\label{fig:bag}
%\vspace{-.1in}
%\end{figure}

\begin{figure}[t]
\centering
\begin{subfigure}[t]{0.49\linewidth}
\centering
  \includegraphics[width=\linewidth]{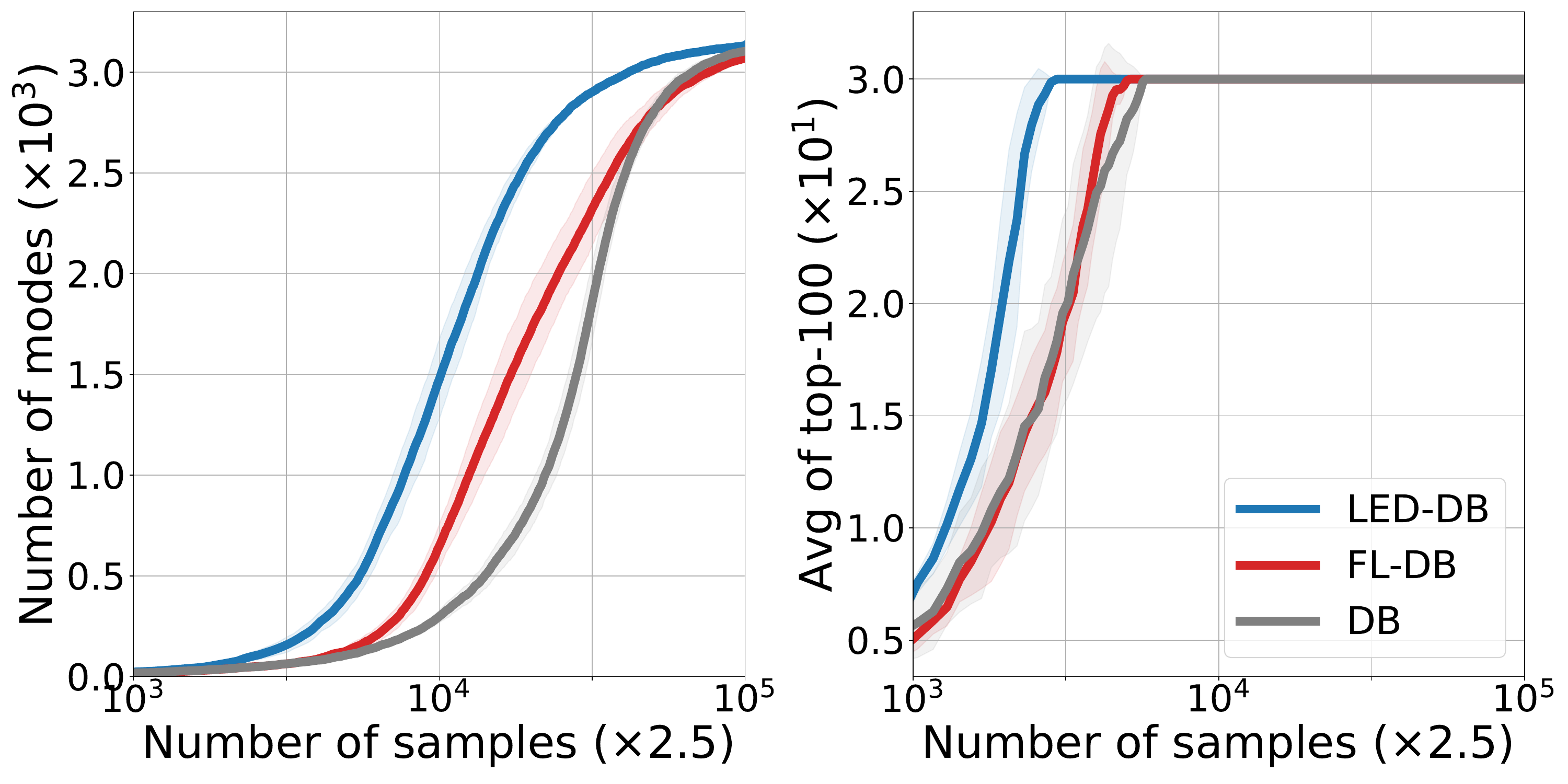}
  \subcaption{DB-based objectives}
\end{subfigure}
\hfill
\begin{subfigure}[t]{0.49\linewidth}
\centering
  \includegraphics[width=\linewidth]{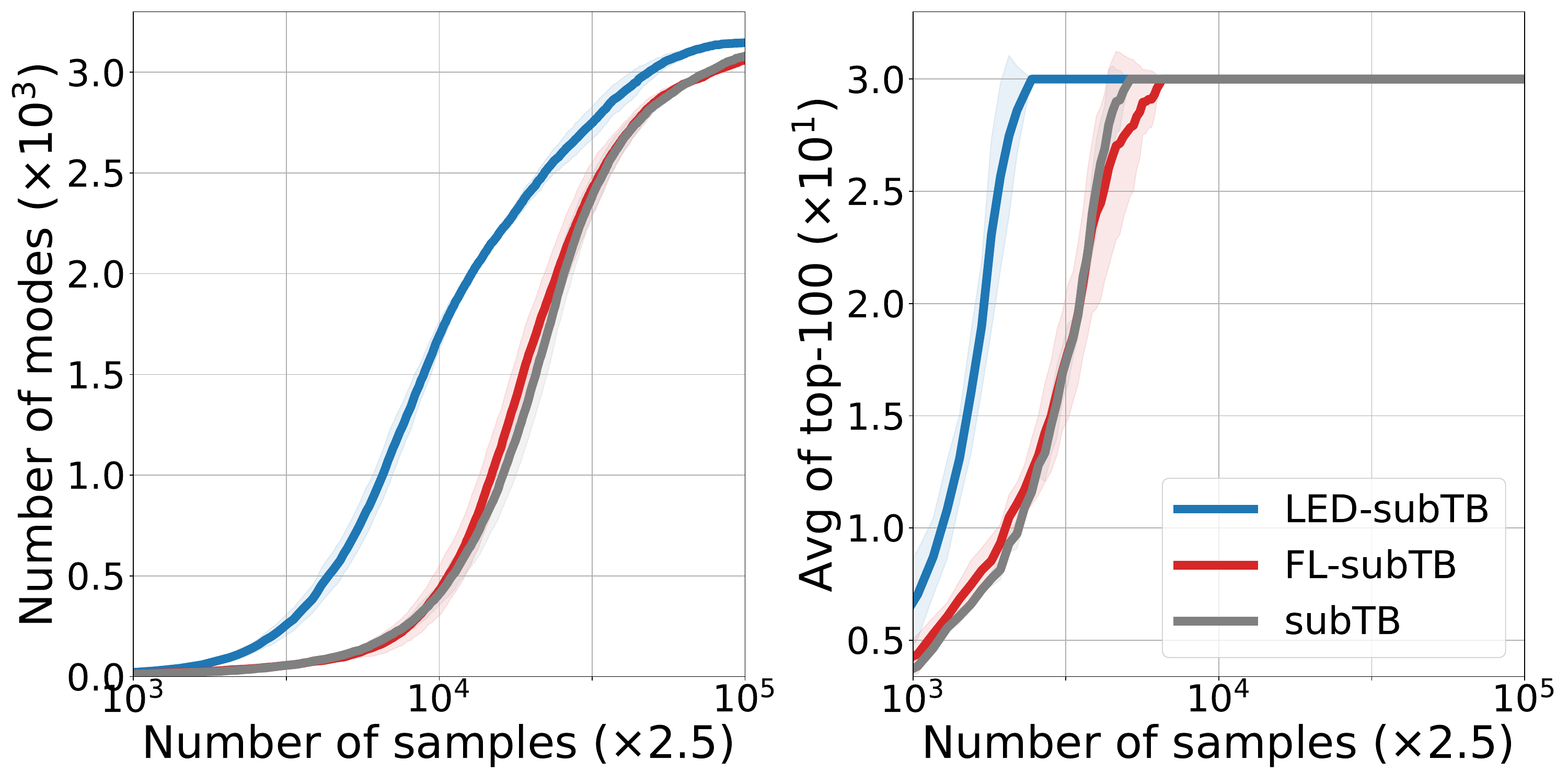}
  \subcaption{SubTB-based objectives}
\end{subfigure}
\caption{
\textbf{The performance on bag generation.} The solid line and shaded region represent the mean and standard deviation, respectively. The LED-GFN shows superiority compared to the considered baselines on both DB-based and subTB-based implementations.
}\label{fig:bag}
\end{figure}

\subsection{Bag generation} \label{subsec:experiment_bag}

First, we consider a bag generation task \citep{towardsunderstandinggflownets}. The action is adding an object from seven distinct entities to a bag with a maximum capacity of $15$. The bag exhibits low energy when including seven or more repeated entities of the same type. In this task, we compare our method with GFN and FL-GFN. We consider both DB-based \citep{bengio2021gflownet} and subTB-based \citep{malkin2022trajectory} implementations. The detailed experimental settings are described in \Cref{appx:exp_setup}.

\textbf{Results. } We present the results in \Cref{fig:bag}. Here, one can observe that our method excels in bag generation compared to GFN and FL-GFN on both DB and subTB. In particular, FL-GFN fails to make improvements on the subTB-based implementation, since most states do not provide informative signals for partial inference (as illustrated in \Cref{fig:first_fig}). In contrast, LED-GFN consistently improves performance by producing informative potentials to enhance partial inference.

\begin{figure}[t]
\centering
\includegraphics[width=0.84\linewidth]{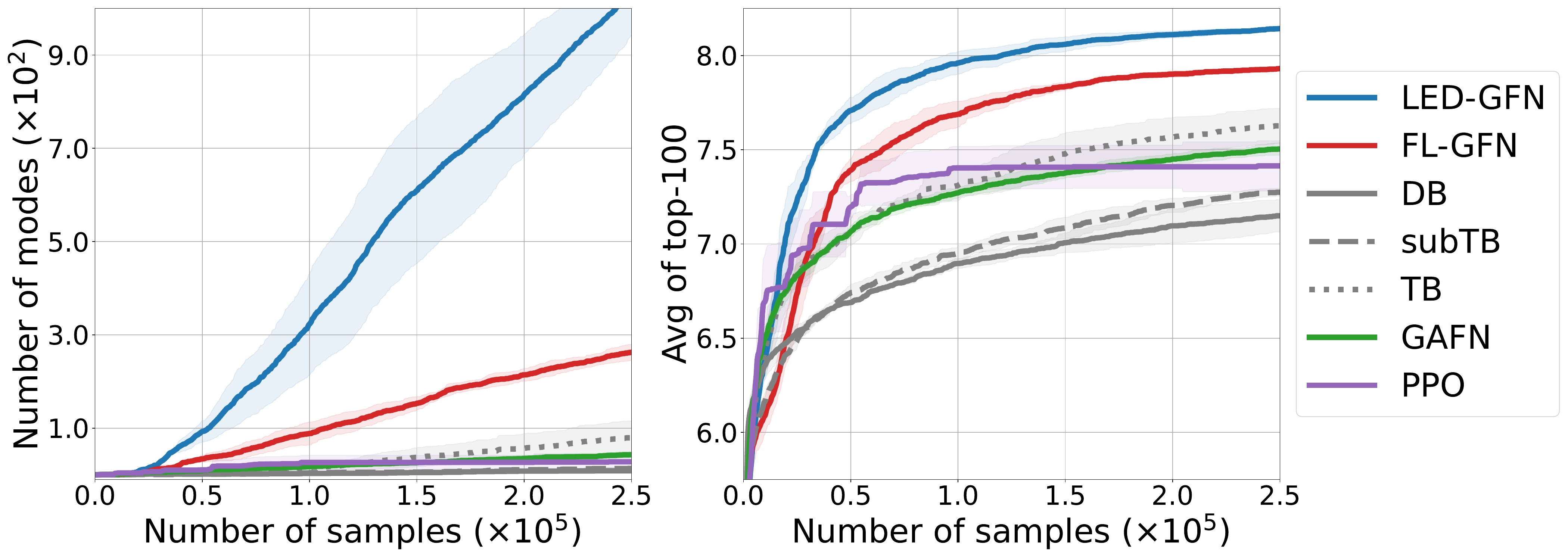}
\caption{\textbf{The performance on molecule generation.} The solid line and shaded region represent the mean and standard deviation, respectively. The LED-GFN shows superiority compared to the considered baselines in generating diverse high reward molecules.
}\label{fig:molecule}
\vspace{-.1in}
\end{figure}

%\begin{figure}[t]
%\centering
%\begin{subfigure}[t]{0.49\linewidth}
%\centering
%  \includegraphics[width=\linewidth]{resources/figures/mols_paper_inter_db.pdf}
%  \subcaption{DB-based objectives}
%\end{subfigure}
%\hfill
%\begin{subfigure}[t]{0.49\linewidth}
%\centering
%  \includegraphics[width=\linewidth]{resources/figures/mols_paper_inter_subtb.pdf}
%  \subcaption{SubTB-based objectives}
%\end{subfigure}
%\caption{
%The molecule generation tasks. The LED-GFN shows superiority compared to the FL-GFlowNet which just use the terminal state energy function to compute intermediate state energy.
%}\label{fig:molecule}
%\end{figure}

\subsection{Molecule generation}

Next, we validate LED-GFN in a more practical domain: the molecule generation task \citep{bengio2021flow}. This task aims to find molecules with low binding energy to the soluble epoxide hydrolase protein. In this task, a molecule is generated by constructing junction trees \citep{jin2018junction}, with the actions of adding molecular building blocks. The binding energy between the molecule and the target protein is computed using a pre-trained oracle \citep{bengio2021flow}.

In this experiment, we consider PPO \citep{schulman2017proximal}, and three GFN models: DB, TB, and subTB \citep{madan2023learning} as the baselines. Additionally, we compare our approach with GAFN \citep{pan2023generative} and FL-GFN. For FL-GFN and LED-GFN, we consider a subTB-based implementation. The overall implementations and experimental settings follow prior studies \citep{bengio2021flow,pan2023better}, which are described in \Cref{appx:exp_setup}.

\textbf{Results. } The results are presented in \Cref{fig:molecule}. One can see that LED-GFN outperforms the considered baselines in enhancing the average score of unique top-$100$ molecules and the number of modes found during training. These results highlight that LED-GFN is also beneficial for real-world generation problems with a large state space.

\subsection{RNA sequence generation}

\begin{figure}[t]
\centering
\includegraphics[width=0.84\linewidth]{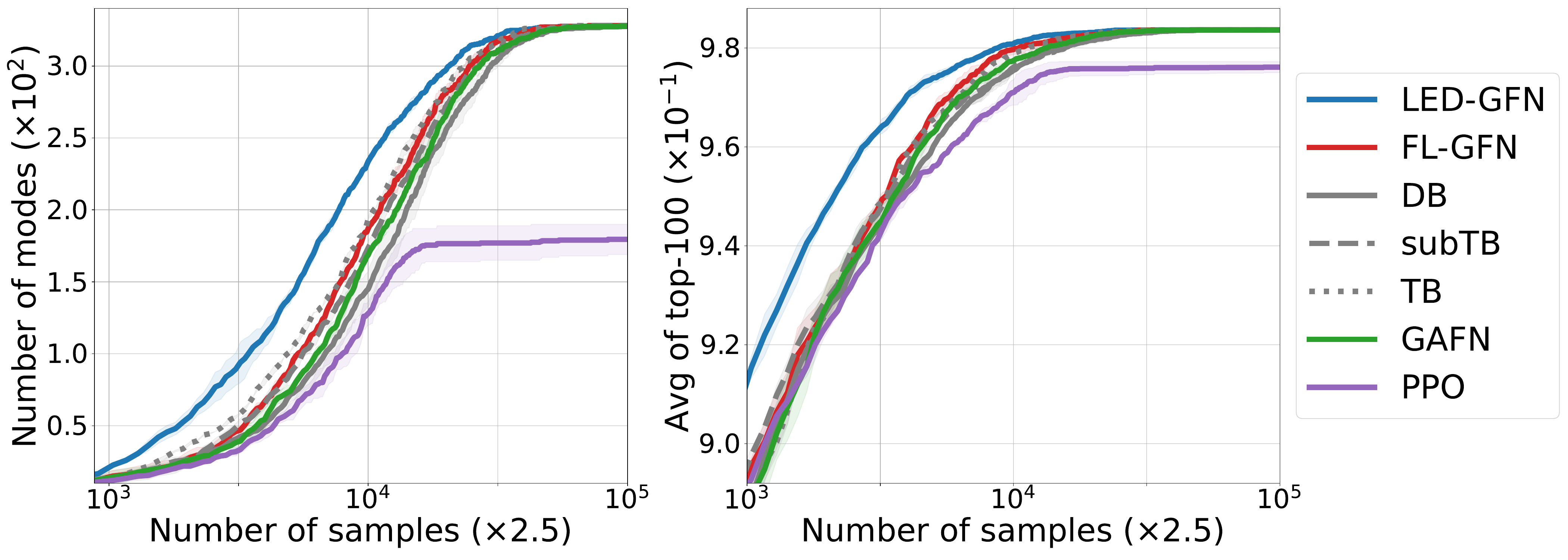}
\caption{\textbf{The performance on RNA sequence generation.} The solid line and shaded region represent the mean and standard deviation, respectively. The LED-GFN shows superiority compared to the considered baselines in generating diverse high reward RNA sequences.
}\label{fig:protein}
\end{figure}

%\begin{figure}[t]
%\centering
%\begin{subfigure}[t]{0.49\linewidth}
%\centering
%  \includegraphics[width=\linewidth]{resources/figures/sub_tfbind_inter_DB.pdf}
%  \subcaption{DB-based objectives}
%\end{subfigure}
%\hfill
%\begin{subfigure}[t]{0.49\linewidth}
%\centering
%  \includegraphics[width=\linewidth]{resources/figures/sub_tfbind_inter_subTB.pdf}
%  \subcaption{SubTB-based objectives}
%\end{subfigure}
%\caption{
%The protein generation tasks. The LED-GFN shows superiority compared to the FL-GFlowNet which just use the terminal state energy function to compute intermediate state energy.
%}\label{fig:protein}
%\end{figure}

We also consider a RNA sequence generation task for discovering diverse and promising RNA sequences that bind to human transcription factors \citep{barrera2016survey,trabucco2022design,jain2022biological}. The action is appending or prepending an amino acid to the current sequence. The energy is pre-computed based on wet-lab measured DNA binding activity to Sine Oculis Homeobox Homolog 6 \citep{barrera2016survey}. We consider the same baselines as in the molecule generation task.

\textbf{Results. } The results are presented in \Cref{fig:protein}. One can observe that LED-GFN outperforms the considered baselines. Furthermore, FL-GFN only makes minor differences compared to GFN, while LED-GFN makes noticeable improvements. These results highlight that energy-based partial inference can fail to improve performance in practical domains, while the potential learning-based approach consistently leads to improvements.

\subsection{Comparison with ideal local credits}
In these experiments, we demonstrate that LED-GFN can achieve similar performance compared to FL-GFN, \textbf{even when the intermediate state energy is sufficient to identify the contribution of the action, i.e., ideal local credit} \citep{zhang2023let}. Note that this tasks are idealized, since designing such an energy function requires a complete understanding of the domain. Especially, we focus on two tasks: set generation \citep{pan2023better} and the maximum independent set problem \citep{zhang2023let}. For these tasks, we compare our method with GFN and FL-GFN. 

\textbf{Set generation.} The set generation task is similar to the bag generation. The actions are adding an objects from $30$ distinct objects to a set with a maximum capacity of $20$. The energy is evaluated by accumulating the individual energy of each entity, so the intermediate energy gain has complete information to identify the contribution of each action \citep{pan2023better}. We describe the detailed experiment settings in \Cref{appx:details_LED}. 

\textbf{Maximum independent set problem.} This task aims to find the maximum independent set by selecting nodes, and the energy is evaluated based on the current size of the independent set \citep{zhang2023let}. Here, we compare the performance on validation graphs following \citet{zhang2023let}. At each step, we sample $50$ solutions for each validation graph to measure the average scores and the number of mode founds (greater than $18.5$). The overall implementations and hyper-parameters follow prior studies \citep{zhang2023let}.

%\begin{figure}[t]
%\centering
%\begin{subfigure}[t]{0.49\linewidth}
%\centering
%  \includegraphics[width=\linewidth]{resources/figures/paper_set_small_inter.pdf}
%  \subcaption{set. Summary of results TOdo: fontsize}
%\end{subfigure}
%\hfill
%\begin{subfigure}[t]{0.49\linewidth}
%\centering
%  \includegraphics[width=0.5\linewidth]{example-image}
%  \subcaption{results on large size set (should be updated)}
%\end{subfigure}
%\caption{
%The set generation tasks. The solid line and shaded region represent the mean and standard deviation of the metric, respectively. The LED-GFN shows similar performance to the FL-GFlowNet, which incorporates dense intermediate energy based on the underlying energy gain mechanism.
%}\label{fig:set}
%\vspace{-.1in}
%\end{figure}

\begin{figure}[t]
\centering
\begin{subfigure}[t]{0.49\linewidth}
\centering
  \includegraphics[width=\linewidth]{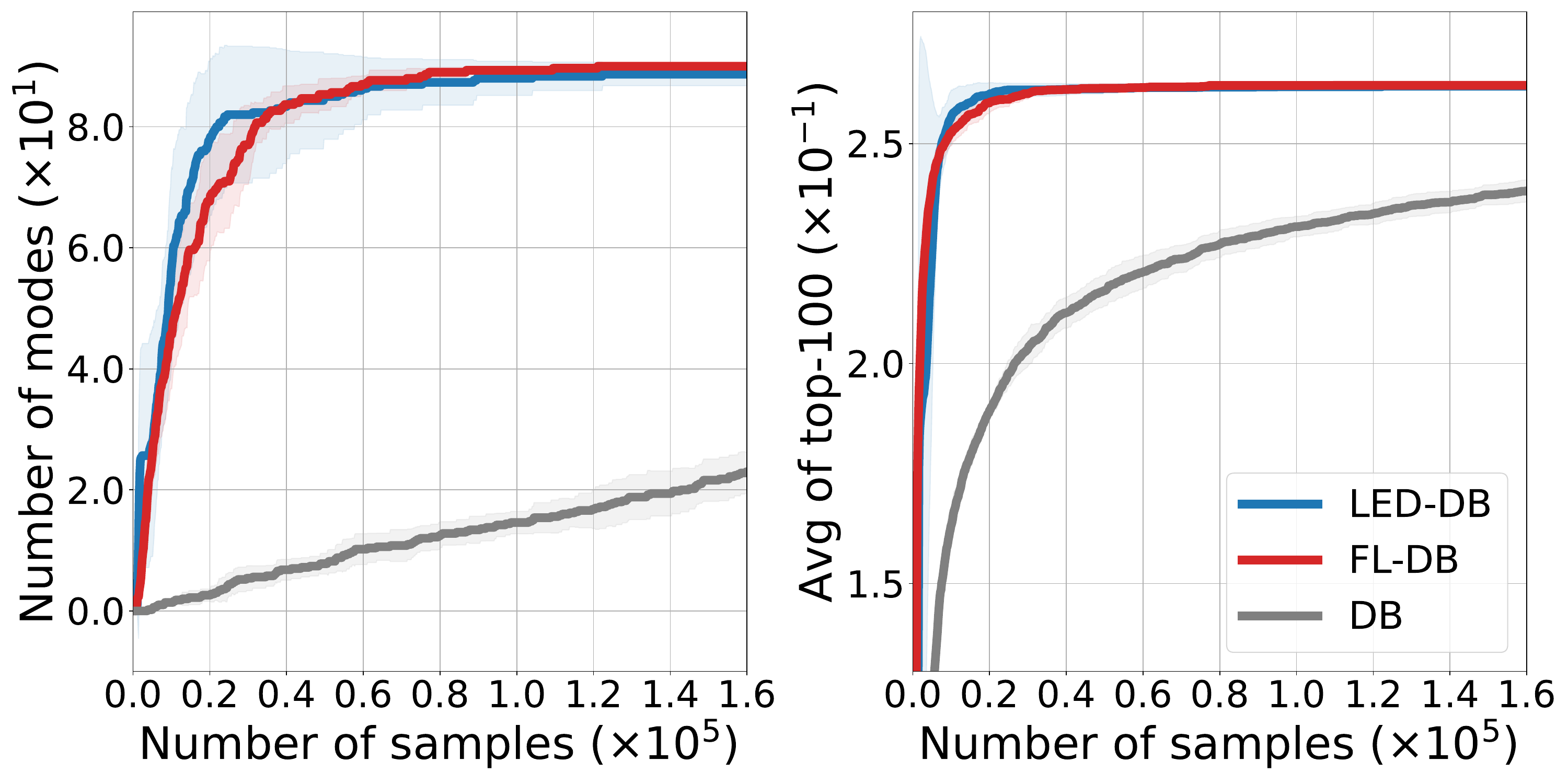}
  \subcaption{Set generation}
\end{subfigure}
\hfill
\begin{subfigure}[t]{0.49\linewidth}
\centering
  \includegraphics[width=\linewidth]{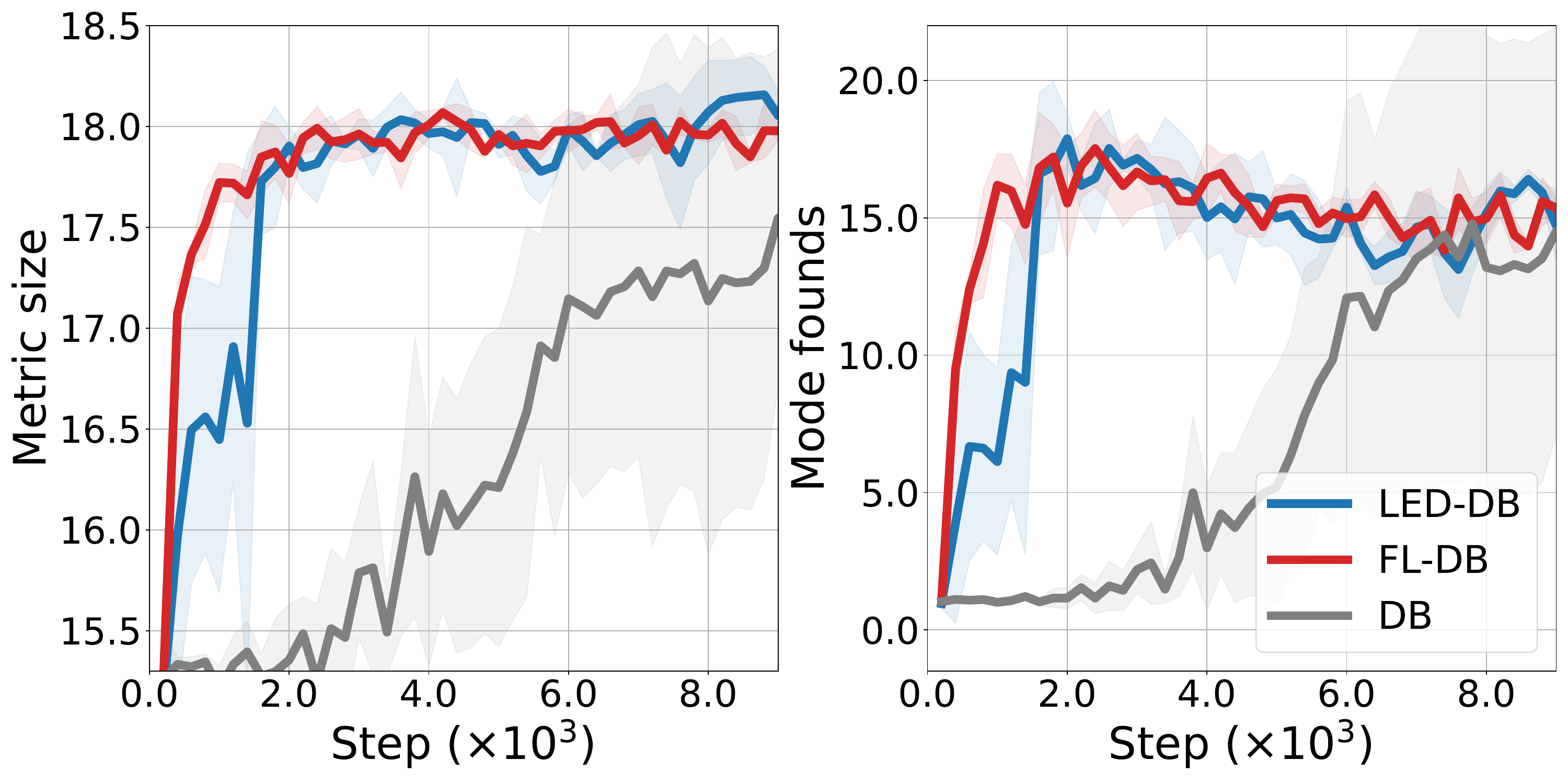}
  \subcaption{Maximum independent set problem}
\end{subfigure}
\caption{
\textbf{The performance comparison with ideal local credits.} The solid line and shaded region represent the mean and standard deviation, respectively. The LED-GFN shows similar performance to the FL-GFN, even when the intermediate energy is sufficient to identify the action contribution.
}\label{fig:set}
\end{figure}

\textbf{Results. } As illustrated in \Cref{fig:set}, one can see that our approach achieves similar performance to FL-GFN, even though the intermediate state energy provides ideal local credit for partial inference. These results highlight that the potentials of LED-GFN can be as informative as ideal local credit, which provides complete identification of the action contributions.

\subsection{Ablation studies} \label{subsec:ablation}

\begin{figure}[t]
\centering
\begin{subfigure}[t]{0.32\linewidth}
\centering
  \includegraphics[width=\linewidth]{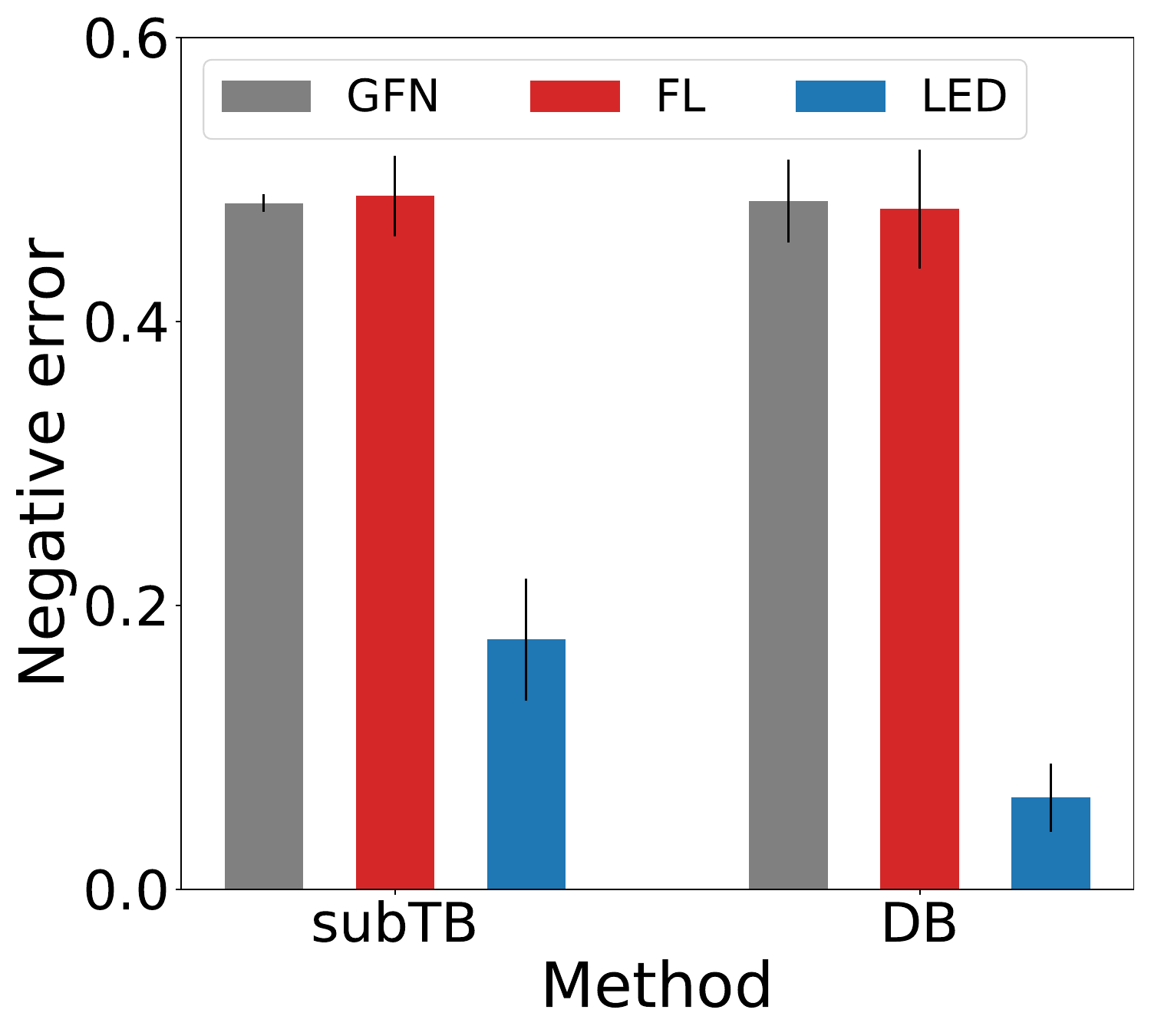}
  \subcaption{Bag generation ($\downarrow$)}\label{subfig:relative_bag}
\end{subfigure}
\hfill
\begin{subfigure}[t]{0.32\linewidth}
\centering
  \includegraphics[width=\linewidth]{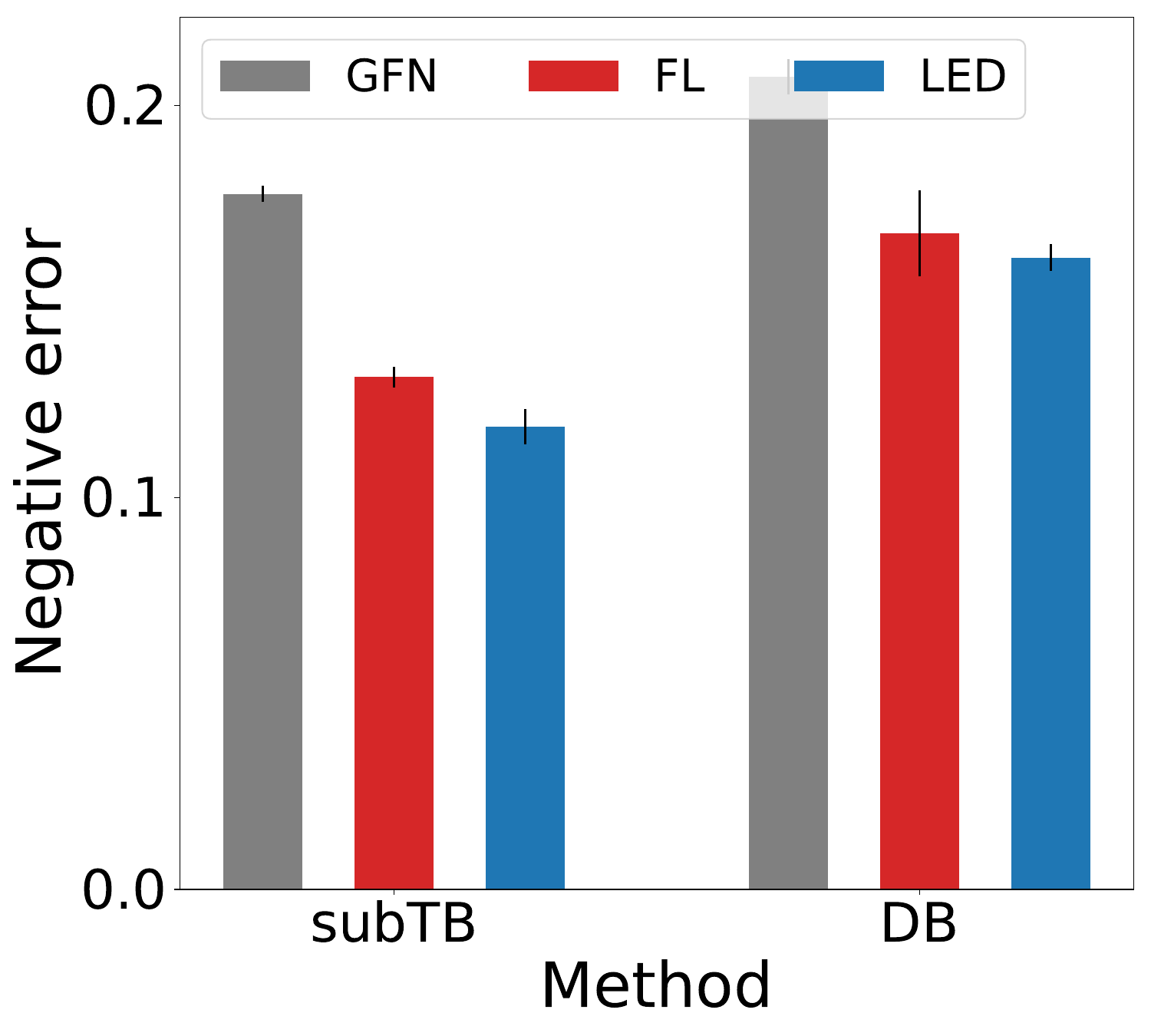}
  \caption{RNA sequence generation ($\downarrow$)}\label{subfig:relative_rna}
\end{subfigure}
\hfill
\begin{subfigure}[t]{0.32\linewidth}
\centering
  \includegraphics[width=\linewidth]{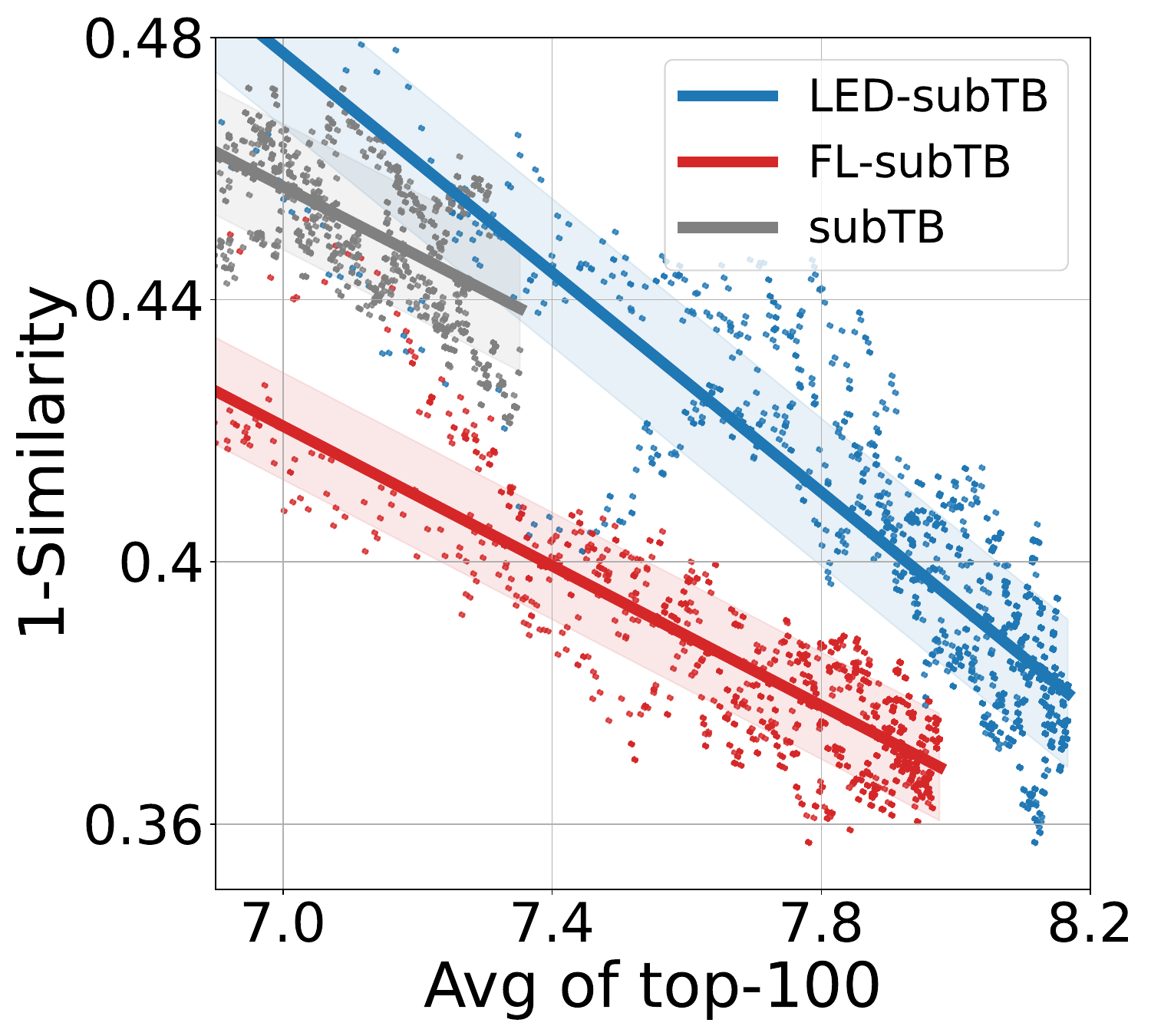}
  \subcaption{Molecule generation ($\shortarrow{1}$)}\label{subfig:simil_mol}
\end{subfigure}
\caption{\textbf{(a-b) The negative relative mean error comparison (lower is better).} Our LED-GFN better approximates the target distributions. \textbf{(c) The Tanimoto similarity with respect to the average scores (upper-right is better).} Our LED-GFN produces more diverse and promising molecules.
}\label{fig:relative}
\end{figure}

\textbf{Goodness-of-fit to the true Boltzmann distribution.} We show that how well our algorithm make a good sampling distribution to sample from the target distribution, i.e., Boltzmann distribution. Following \citet{towardsunderstandinggflownets}, we measure the relative mean error between the empirical generative distribution and target distribution following. In \cref{subfig:relative_bag,subfig:relative_rna}, one can observe that LED-GFN achieves a better approximation to target distribution compared to considered baselines.

\textbf{Diversity vs. high score.} Next, we further verify that our algorithm not only generates high-scoring samples but also diverse molecules. Specifically, we analyze the trade-off between the average score of the top-$100$ samples and the diversity these samples. To measure diversity, we compute the average pairwise Tanimoto similarity \citep{bajusz2015tanimoto}. In Figure \ref{subfig:simil_mol}, we illustrate the Tanimoto similarities with respect to the average of top-$100$ scores. Here, one can observe that LED-GFN achieves better diversity with respect to the average scores.  

\begin{figure}[t]
\centering
\begin{subfigure}[t]{0.49\linewidth}
\centering
  \includegraphics[width=\linewidth]{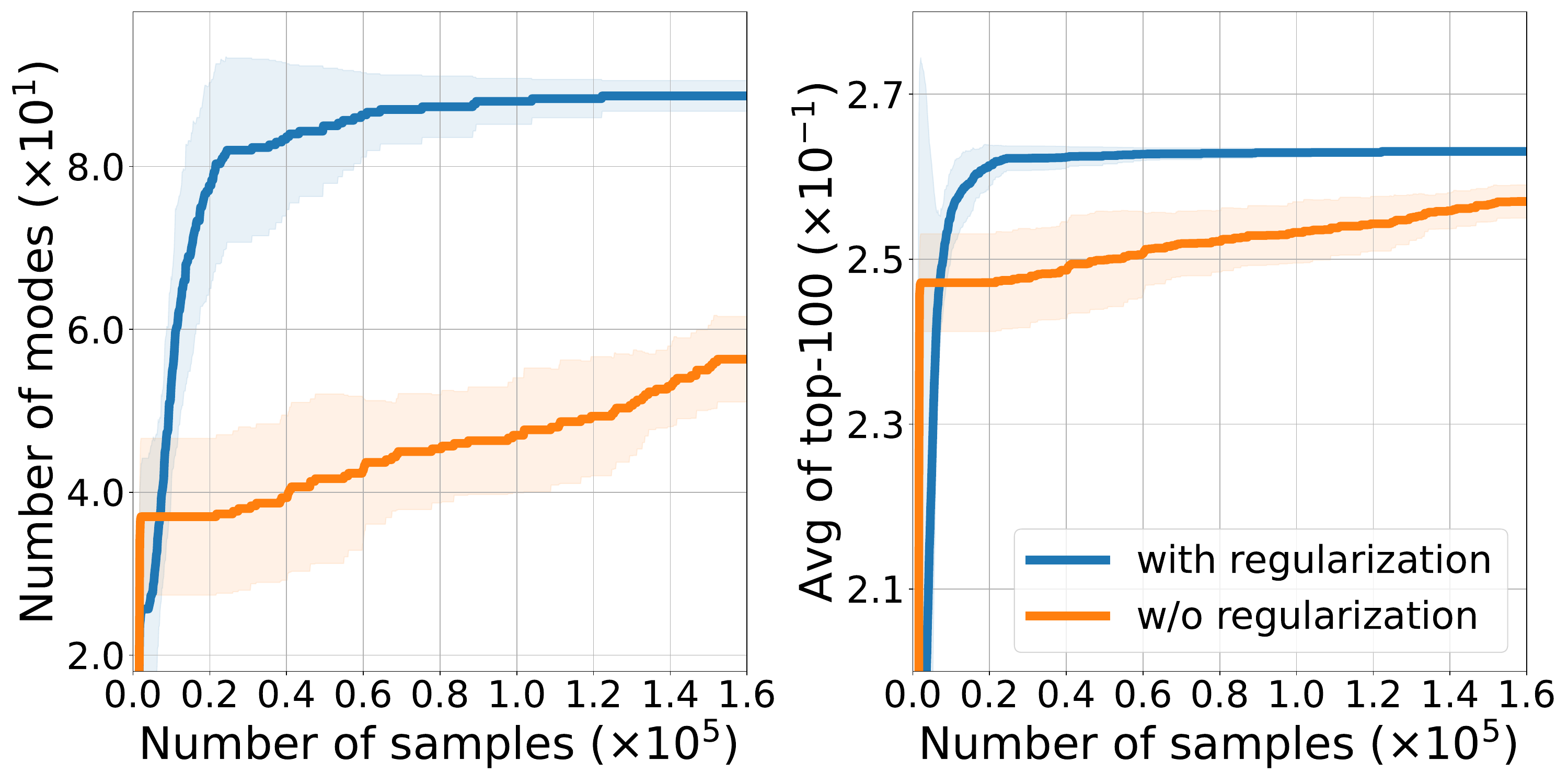}
  \subcaption{Set generation}\label{fig:regularization}
      \vspace{-.03in}
\end{subfigure}
\hfill
\begin{subfigure}[t]{0.49\linewidth}
\centering
  \includegraphics[width=\linewidth]{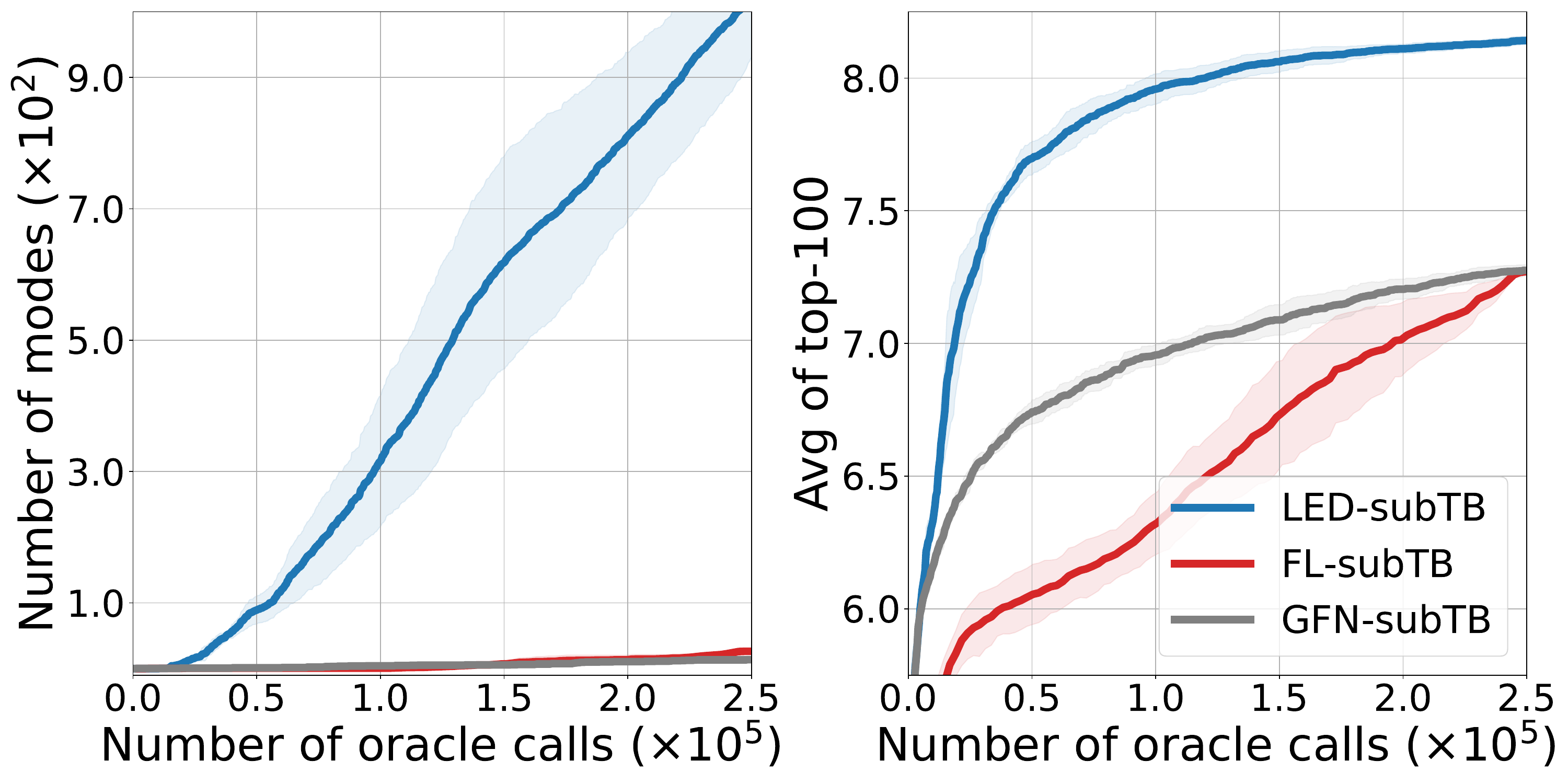}
  \subcaption{Molecule generation}\label{fig:oracle_calls}
      \vspace{-.03in}
\end{subfigure}
\caption{\textbf{(a) The benefits of regularizing variance of potentials.} The regularization improves the performance. \textbf{(b) The performance over the number of energy evaluation.} The FL-GFN can not make improvement with respect to the number of energy evaluation. 
}
\end{figure}

%\begin{wrapfigure}{r}{0.5\textwidth}
%\centering
%\vspace{-.1in}
%    \includegraphics[width=0.95\linewidth]{resources/figures/paper_set_small_model_wo.pdf}
%    \vspace{-.1in}
%    \caption{\textbf{The effects of reducing variance.} }\label{fig:reducing_variance}
%    \vspace{-.1in}
%\end{wrapfigure}

\textbf{Regularized vs. non-regularized potentials.} We also analyze how reducing the variance of potentials benefits the improvement in performance. To this end, we compare energy decomposition learning with variance regularization and its counterpart without regularization, i.e., set $\lambda=1$ in \Cref{eq:least_square}. In \Cref{fig:regularization}, one can see that applying regularization yields more promising performance. These results highlight that inducing dense potentials plays a significant role in improving performance.

\textbf{Number of energy evaluation vs. performance} We analyze the performance with respect to the number of energy evaluation, which can be expensive. In \Cref{fig:oracle_calls}, one can see that FL-GFN can not improve performance since it requires evaluating the energy for every visited states. In contrast, LED-GFN uses a potential function without energy evaluation for intermediate states.

\begin{figure}[t]
\centering
\begin{subfigure}[t]{0.49\linewidth}
\centering
  \includegraphics[width=\linewidth]{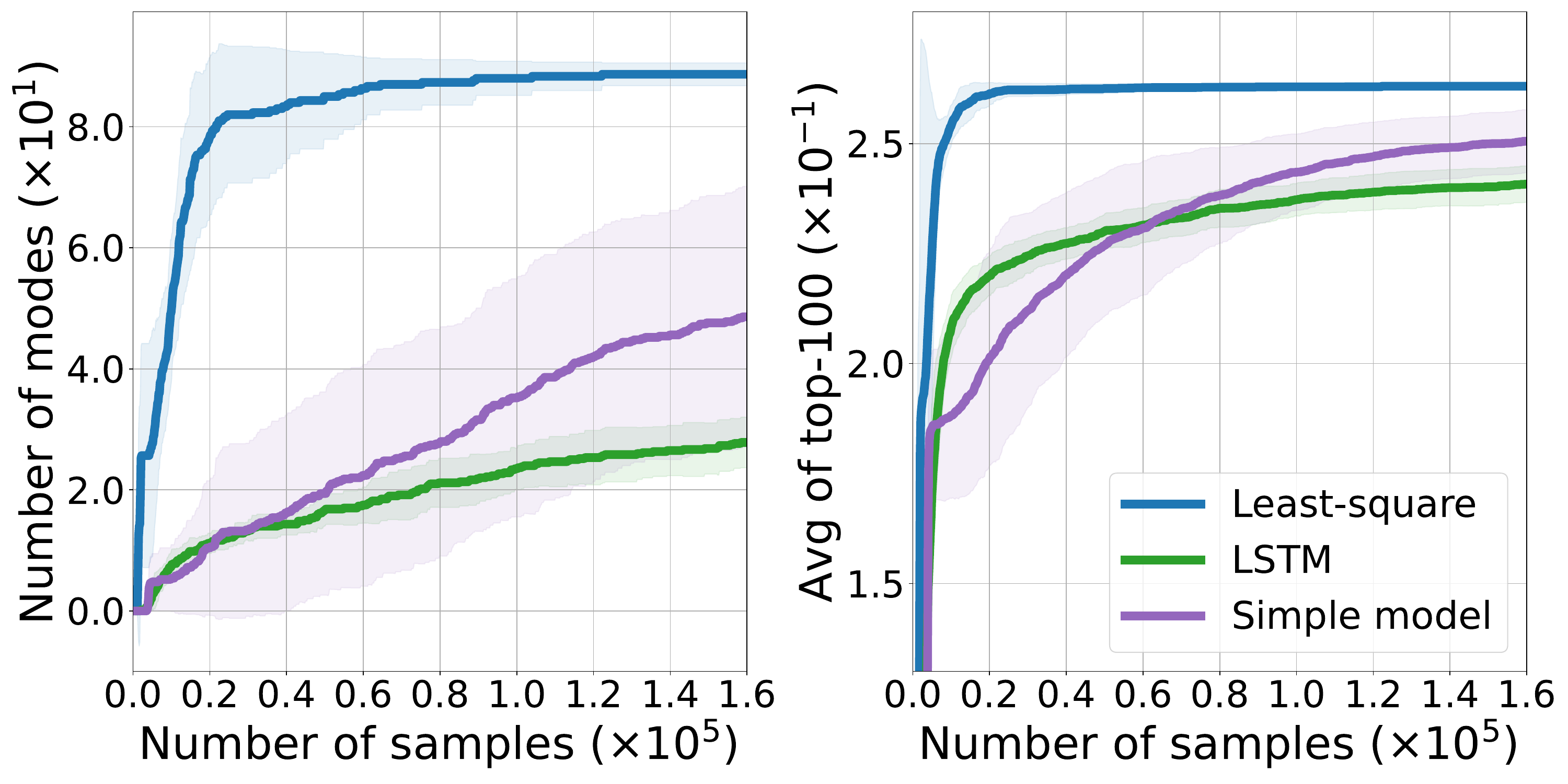}
  \subcaption{Set generation}
      \vspace{-.03in}
\end{subfigure}
\hfill
\begin{subfigure}[t]{0.49\linewidth}
\centering
  \includegraphics[width=\linewidth]{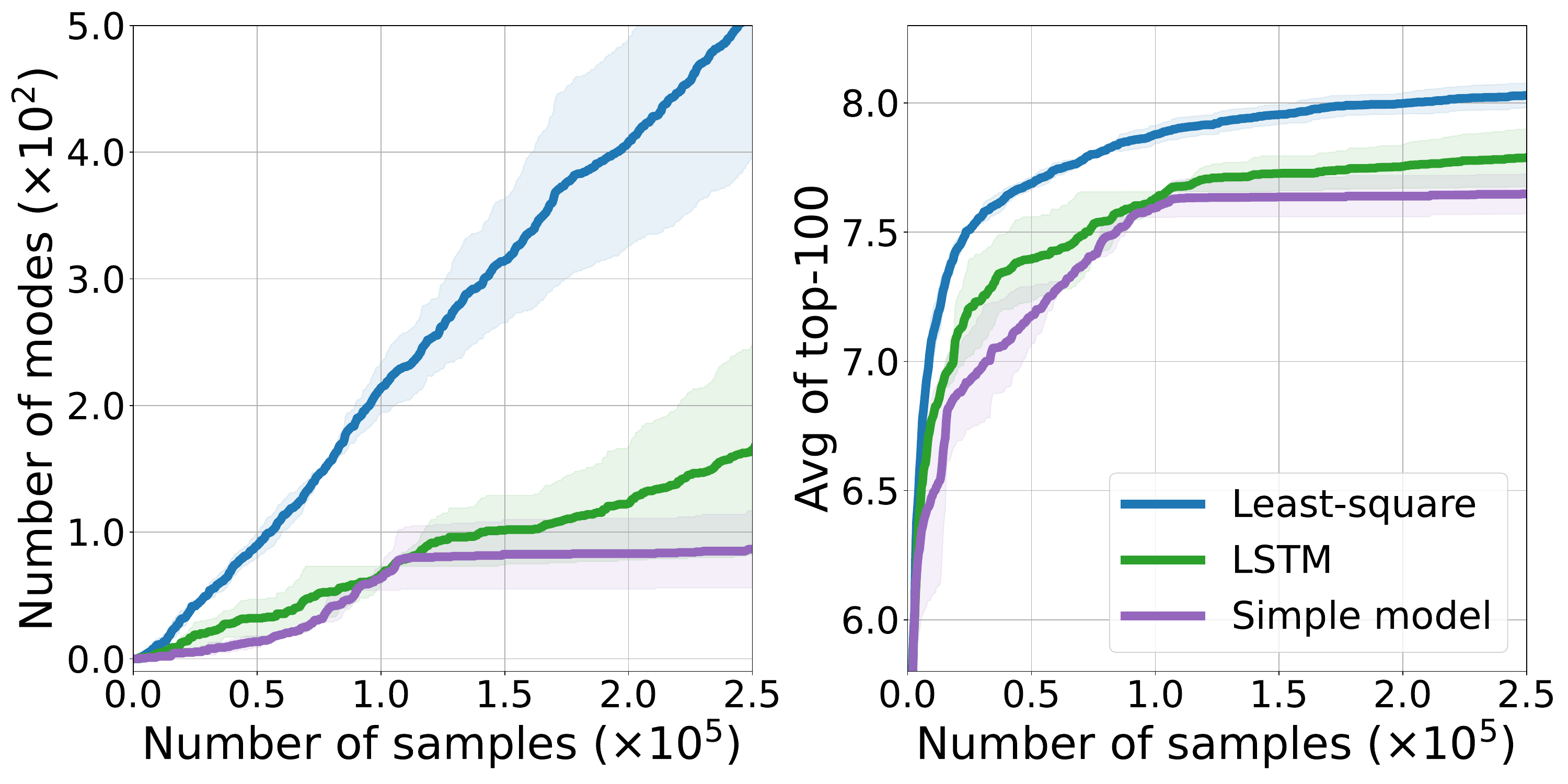}
    \subcaption{Molecule generation}
    \vspace{-.03in}
\end{subfigure}
\caption{\textbf{Comparison between potential learning methods for partial inference.} The least-square based energy decomposition shows most promising results for partial inference.
}\label{fig:method_compare}
\vspace{-.1in}
\end{figure}

\textbf{Energy decomposition learning vs. alternative potentials learning.} We further investigate the following alternative potential learning schemes.
\begin{itemize}[topsep=-1.0pt,itemsep=1.0pt,leftmargin=3.0mm]
\item One may train a proxy model $\phi_{\theta}:\mathcal{X}\rightarrow \mathbb{R}$ to predict the terminal energy, and utilize it to compute potentials $\phi_{\theta}(s\rightarrow s')=\phi_{\theta}(s')-\phi_{\theta}(s)$. This approach can be interpreted as extension of model-based GFlowNet \citep{jain2022biological} for partial inference.
\item Based on the LSTM-based decomposition method \citep{arjona2019rudder}, one can design the potential $\phi_{\theta}(s_t \rightarrow s_{t+1})$ as the difference between two subsequent predictions for $(s_0, a_0, \ldots, s_t, a_t)$ and $(s_0, a_0, \ldots, s_{t+1}, a_{t+1})$ using an LSTM. %one can consider a LSTM which takes trajectory $(s_0,a_0,\cdots,s_{n-1},a_{n-1})$ to predict the object energy $\mathcal{E}(x)$. Then, the potential $\phi_{\theta}(s_t\rightarrow s_{t+1})$ can be designed as the difference between two subsequent predictions on $(s_0,a_0,\cdots,s_{t},a_{t})$ and $(s_0,a_0,\cdots,s_{t+1},a_{t+1})$.
\end{itemize}

In \Cref{fig:method_compare}, we compare each method in molecule and set generation tasks with a DB-based implementation. Here, one can see that the least square-based approach shows the most competitive performance due to its capabilities in producing dense and informative potentials.

%\textbf{Analysis on potentials.} We also analyze how potential function evaluates local credit. In \Cref{fig:first_fig}, one can see that LED-GFN produces informative local credit. We further analyze the 

\section{Conclusion}
\label{sec:conclusion}

In this paper, we propose learning energy decomposition for GFlowNets (LED-GFN). Experiments on various domains show that LED-GFN is promising compared to existing partial inference methods for GFlowNet. An interesting avenue for future work is developing new partial inference techniques using learnable local credit, other than the flow reparameterization considered in our work. 

%An interesting avenue for future work is analyzing the gradient variance of partial inference-based frameworks \citep{madan2023learning}, or shaping local credit for improved exploration and exploitation \citep{sun2022exploit}. These can be new research directions for designing local credit to enhance GFlowNets.

\newpage

\textbf{Reproducibility.} We describe experimental details in \Cref{appx:exp_setup}, which provides the base implementation references, environments, and detailed hyper-parameters. In the supplementary materials, we also include the codes for molecule generation tasks based on the official implementation codes of the prior study \citep{pan2023better}.

\bibliography{ref.bib}
\bibliographystyle{iclr2024_conference}

\appendix
\onecolumn

%\section{Appendix}
%\label{sec:appendix}

\section{Related works} \label{appx:related_work}

\textbf{Generative augmented flow network \citep[GAFN]{pan2023generative}.} The GAFN is a learning framework that incorporates intermediate rewards for exploration purposes. Specifically, GAFN measures the novelty score of a given state, which provides an intrinsic signal to facilitate better exploration towards unvisited states. To compute the novelty score, this method also incorporates online training of random network distillation \citep{burda2018exploration}, which assigns lower scores to unseen states compared to more frequently observed states.

\textbf{Model-based GFlowNet.} \citet{jain2022biological} propose model-based training of GFlowNet for discovering diverse and promising biological sequences. They train a proxy model of the energy function to mitigate the expensive cost of evaluating biological sequences, such as wet-lab evaluation. Additionally, they introduce an active learning algorithm for the model-based GFlowNet, leveraging the epistemic uncertainty estimation of the model to improve exploration.

\textbf{Return decomposition learning.} Our LED-GFN approach is inspired by return decomposition learning of reinforcement learning in sparse reward settings. Their goal is to decompose the return into step-wise dense reward signals \citep{arjona2019rudder, gangwani2020learning, ren2022learning}. They have studied various return decomposition methods. First, \citet{arjona2019rudder} utilize an LSTM-based model to produce step-wise proxy rewards. Next, \citet{gangwani2020learning} propose a simple approach that uniformly redistributes the terminal reward over the trajectory. \citet{ren2022learning} train a proxy reward function using randomized return decomposition learning which is contrained to produces dense and informative proxy rewards.

\newpage

\section{Details of LED-GFN} \label{appx:details_LED}

\subsection{Preserving optimal policy of GFlowNet} \label{subsec:preserving_optimal}

Although the potential function is inaccurate, we show that optimum of $\mathcal{L}_{\text{LED}}$ can induce an optimal policy that samples from Boltzmann distribution. We give a simple proof by reduction to the optimum of DB objective \citep{bengio2021gflownet}. Suppose the parameterization $\phi_{\theta}(s\rightarrow s')=\phi_{\theta}(s')-\phi_{\theta}(s)$, and $\log \hat{F}(s)=-\phi_{\theta}(s)+\log \tilde{F}(s)$. Then, we can reformulate $\mathcal{L}_{\text{LED}}$ as follows: 
\begin{equation*}
\mathcal{L}_{\text{LED}}(s, s') = \left(\log \hat{F}(s)+ \log P_{F}(s' | s) - \log \hat{F}(s')-P_B(s | s')\right)^2.
\end{equation*}
If we add an additional correction term $-\mathcal{E}(x) + \Phi_{\theta}(\tau) = -\mathcal{E}(x) + \phi_{\theta}(x)$ to the terminal flow such that $\log\hat{F}(x) = -\mathcal{E}(x),$ this objective becomes to be equivalent to the reparameterization of DB, where the optimum induces a policy sampling from a Boltzmann distribution \citep{bengio2021gflownet}. Therefore, the optimum of LED-GFN can still induce the policy that samples from the Boltzmann distribution. We refer to this correction-based approach as LED-GFN$^{*}$.

\begin{figure}[h]
\centering
  \includegraphics[width=0.5\linewidth]{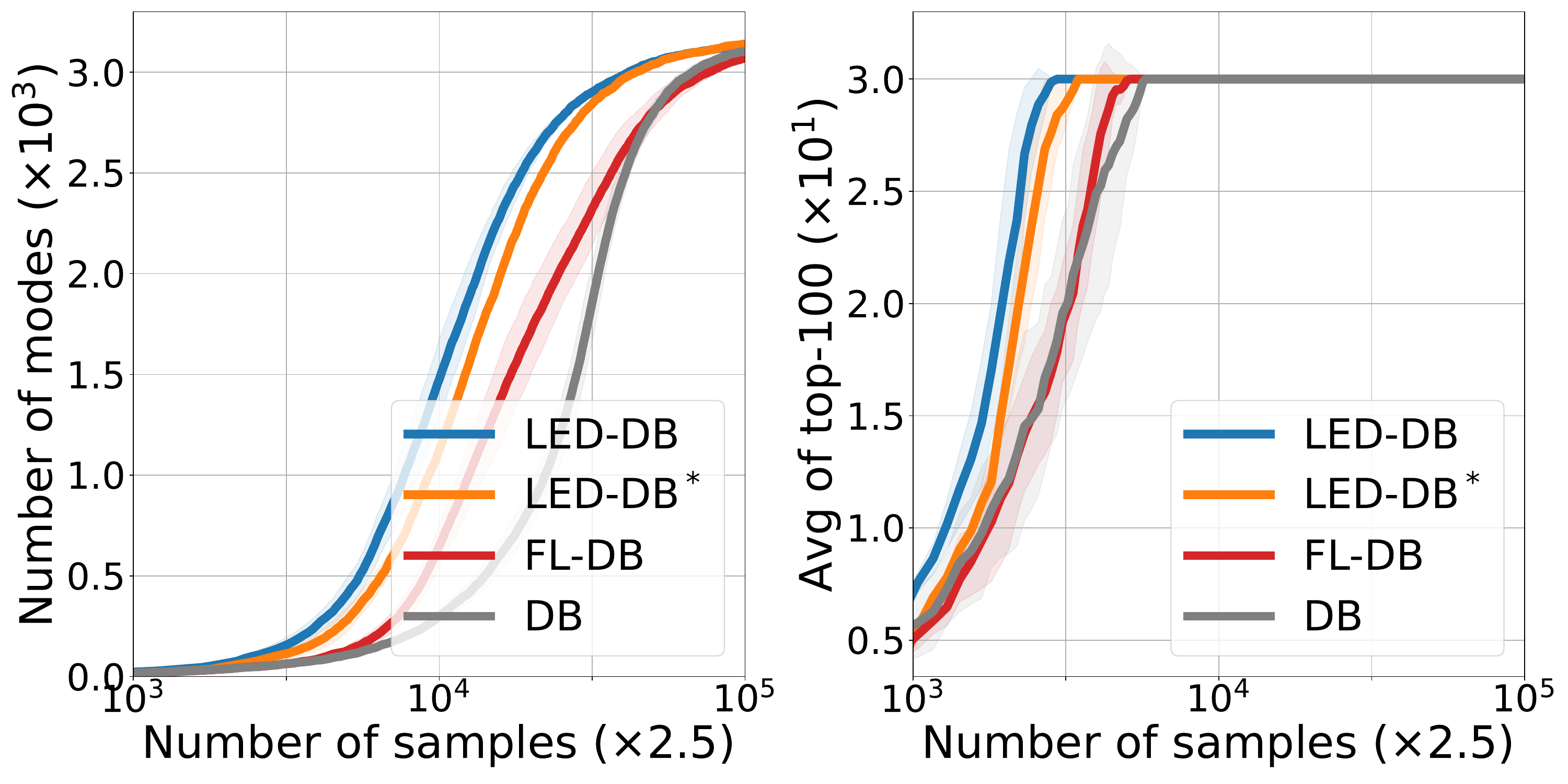}
\caption{The performance on bag generation.
}\label{fig:bag_opt}
\end{figure}

However, our implementation follows a prior study in return decomposition learning \citep{ren2022learning}, which uniformly redistributes the decomposition error over the transitions within the given trajectory (we denote this approach as LED-GFN in experiments). In \Cref{fig:bag_opt}, we empirically observe that this approach further improves the training of GFlowNets. We assume that uniformly redistributed decomposition error partially provides more dense and informative local credit signals correlated with future energy.

\subsection{Training on subTB} \label{subsec:appdix_subtb}

The LED-GFN can also be implemented on subTB by incorporating the potentials within sub-trajectories. To this specific, one can modify subTB as follows:
\begin{align*}
&\mathcal{L}_{\text{LED-subTB}}=\\
&\left(\log \tilde{F}(s_U) +\sum^{V-1}_{t=U}\log P_{F}(s_{t+1}|s_{t})+\sum^{V-1}_{t=U}\phi_{\theta}(s_t\rightarrow s_{t+1})-\log \tilde{F}(s_V)-\sum^{V-1}_{t=U}\log P_B(s_{t}|s_{t+1})\right)^2,
\end{align*}
which is based on a sub-trajectory $s_{U}\rightarrow s_{U+1} \cdots\rightarrow s_{V}$. This objective is equivalent to FL-GFN on the subTB when $\phi_{\theta}(s\rightarrow s') = \mathcal{E}(s') - \mathcal{E}(s)$ \citep{pan2023better}, but we replace it with the potentials for better credit assignment.

\newpage

\section{Experimental details} \label{appx:exp_setup}

For all experiments, the neural network architecture of potential function is identical to that of GFlowNet policy. We set the learning rate for potential functions to $0.001$. The dropout is applied to potentials in least square-based energy decomposition learning \Cref{eq:least_square}. We set the dropout probability as $10\%$ for tasks with a trajectory length less than $10$ and $20\%$ for others.

\textbf{Bag generation \citep{towardsunderstandinggflownets}.} The experiment settings, implementations, and hyper-parameters are based on prior studies \citep{towardsunderstandinggflownets}. The bag generation task is to generate a bag with a maximum capacity of $15$. There are seven types of entities, where each action includes one of them in the current bag. If it contains seven or more repeats of any items, it has reward $10$ with $75\%$ chance, and 30 otherwise. The threshold for determining the mode is $30$. 

In each round, we generate $B_1=32$ bags from the policy. The GFlowNet model consists of two hidden layers with $16$ hidden dimensions, which is trained with a learning rate of $1\mathrm{e}{-4}$. We use an exploration epsilon of $0.01$. In addition, their implementation involves a buffer for enabling backward policy-based learning \citep{towardsunderstandinggflownets,malkin2022gflownets}. We utilize this buffer for energy decomposition learning. The mini-batch size $B_2$ is same as $B_1$. We set the number of iterations in energy decomposition learning $N=8$ for each round. Note that reducing $N$ still leads to promising results compared to baseline.

\textbf{Molecule generation \citep{bengio2021flow}.} The experiment settings, implementations, and hyper-parameters are based on prior studies \citep{bengio2021flow,pan2023better}. The maximum trajectory length is $8$, with the number of actions varying between around $100$ and $2000$ which is depending on the state. The threshold for determining the mode is $7.5$. 

In each round, we generate four molecules, i.e., $B_1=4$. The model consists of Message Passing Neural Networks \citep{gilmer2017neural} with ten convolution steps and $256$ hidden dimensions, which is trained with a learning rate of $5\mathrm{e}{-4}$. We rescale the reward so that the maximum reward is close to one, and exponent of it is set to $8.0$. We use an exploration epsilon of $0.05$. In energy decomposition learning, we do not use a buffer and immediately use the molecules that are sampled in each round. For PPO, we set the entropy coefficient to $1\mathrm{e}{-4}$ and do not apply the reward exponent because it causes a gradient exploding.

\textbf{RNA sequence generation \citep{towardsunderstandinggflownets}.} The experiment settings, implementations, and hyper-parameters are based on prior studies \citep{towardsunderstandinggflownets}. The action is defined as prepending or appending an amino acid to the current sequence. The maximum length is $8$ and the number of actions is $4$. The mode is determined based on whether it is included in a predefined set of promising RNA sequences \citep{towardsunderstandinggflownets}.

In each round, we generate $B_1=16$ sequences. The GFlowNet model consists of two hidden layers with $128$ hidden dimensions, which is trained with a learning rate of $1\mathrm{e}{-4}$. The reward exponent is set to $3.0$. We use an exploration epsilon of $0.01$. In addition, their implementation involves a buffer for enabling backward policy-based learning iteration \citep{towardsunderstandinggflownets}. The hyper-parameters for energy decomposition learning is the same as that of bag generation. For PPO, we set the entropy coefficient to $1\mathrm{e}{-2}$.

\textbf{Set generation \citep{pan2023better}.} The experiment settings, implementations, and hyper-parameters are based on prior studies \citep{pan2023better}. The set generation task is to generate a set which involves $20$ entities. There are $30$ types of entities, where each action includes one of them in the current set. We set the threshold for determining the mode as $0.25$. 

In each round, we generate $B_1=16$ sets. The GFlowNet model consists of two hidden layers with $256$ hidden dimensions, which is trained with a learning rate of $0.001$. The hyper-parameters for energy decomposition learning is the same as that of molecule generation.

\end{document}